\PassOptionsToPackage{usenames,dvipsnames}{xcolor}

\documentclass[10pt,twocolumn,letterpaper]{article}

\usepackage{cvpr}              %

\makeatletter
\@namedef{ver@everyshi.sty}{}
\makeatother
\usepackage{tcolorbox}
\newtcolorbox{mybox}{size=title}

\usepackage{graphicx}
\usepackage{amsmath}
\usepackage{amssymb}
\usepackage{booktabs}

\usepackage[sort&compress,numbers]{natbib}
\usepackage[usenames,dvipsnames]{xcolor}
\usepackage{pifont}
\newcommand{\xmark}{\ding{55}}%
\usepackage{adjustbox}
\usepackage{diagbox}
\usepackage{balance}
\usepackage[utf8]{inputenc}
\usepackage{multirow}
\usepackage{CJKutf8}
\usepackage{siunitx}
\usepackage{arabtex}
\usepackage{utf8}
\setcode{utf8}
\usepackage[symbol]{footmisc}
\usepackage{enumitem}
\usepackage{microtype}
\usepackage{url}
\usepackage[skip=3pt]{caption}
\usepackage{subcaption}

\usepackage[symbol]{footmisc}

\usepackage{titletoc}
\usepackage[accsupp]{axessibility}  %

\DeclareRobustCommand{\mole}{%
  \begingroup\normalfont
  \includegraphics[height=1.5\fontcharht\font`\F]{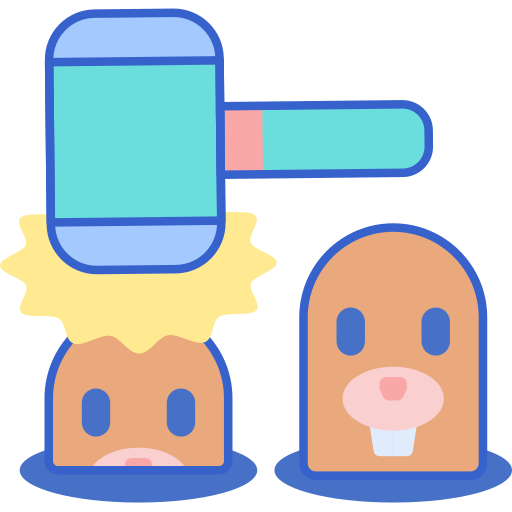}%
  \endgroup
}

\DeclareRobustCommand{\molehammer}{%
  \begingroup\normalfont
  \includegraphics[height=1.5\fontcharht\font`\F]{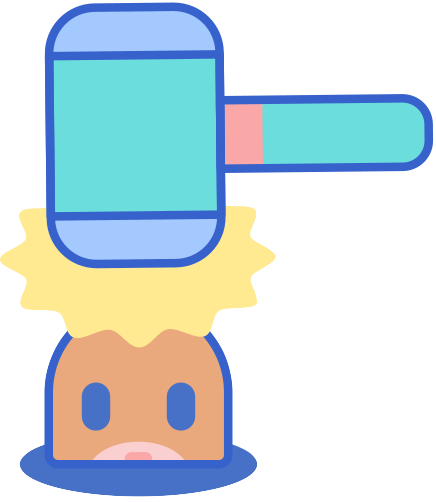}%
  \endgroup
}

\DeclareRobustCommand{\molenohammer}{%
  \begingroup\normalfont
  \includegraphics[height=1.5\fontcharht\font`\F]{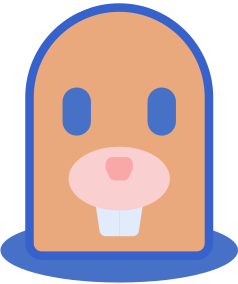}%
  \endgroup
}

\definecolor{citecolor}{HTML}{0071bc}

\usepackage{colortbl}
\usepackage{xcolor}
\definecolor{lightgray}{gray}{0.5}

\usepackage[pagebackref=true,breaklinks=true,colorlinks,citecolor=citecolor,bookmarks=false]{hyperref}

\usepackage[capitalize]{cleveref}
\crefname{section}{Sec.}{Secs.}
\Crefname{section}{Section}{Sections}
\Crefname{table}{Table}{Tables}
\crefname{table}{Tab.}{Tabs.}

\makeatletter
\def\@fnsymbol#1{\ensuremath{\ifcase#1\or \dagger\or \ddagger\or
   \mathsection\or \mathparagraph\or \|\or **\or \dagger\dagger
   \or \ddagger\ddagger \else\@ctrerr\fi}}
\makeatother

\makeatletter
\def\NAT@def@citea{\def\@citea{\NAT@separator}}
\makeatother

\def\cvprPaperID{***} %
\def\confName{CVPR}
\def\confYear{2023}

\begin{document}

\title{
\vspace{-11mm} A Whac-A-Mole Dilemma \mole:\\
Shortcuts Come in Multiples Where Mitigating One \molehammer{} Amplifies Others \molenohammer  \vspace{-6mm}
}

\author{
\thanks{Work done during the internship at Meta AI. $^*$Equal Contribution.}~Zhiheng Li$^{2}$
\quad $^*$Ivan Evtimov$^{1}$
\quad Albert Gordo$^{1}$
\quad Caner Hazirbas$^{1}$
\quad Tal Hassner$^{1}$ \\
Cristian Canton Ferrer$^{1}$
\quad Chenliang Xu$^{2}$
\quad $^*$Mark Ibrahim$^{1}$ \\[0.5mm]
$^{1}$Meta AI \quad $^{2}$University of Rochester \\
{\tt\small \{ivanevtimov,agordo,hazirbas,thassner,ccanton,marksibrahim\}@meta.com} \\
{\tt\small \{zhiheng.li,chenliang.xu\}@rochester.edu} \vspace{-5mm}
}
\maketitle

\begin{abstract}
Machine learning models have been found to learn shortcuts---unintended decision rules that are unable to generalize---undermining models' reliability. Previous works address this problem under the tenuous assumption that only a single shortcut exists in the training data. Real-world images are rife with multiple visual cues from background to texture. Key to advancing the reliability of vision systems is understanding whether existing methods can overcome multiple shortcuts or struggle in a Whac-A-Mole game, i.e., where mitigating one shortcut amplifies reliance on others. To address this shortcoming, we propose two benchmarks: 1) UrbanCars, a dataset with precisely controlled spurious cues, and 2) ImageNet-W, an evaluation set based on ImageNet for watermark, a shortcut we discovered affects nearly every modern vision model. Along with texture and background, ImageNet-W allows us to study multiple shortcuts emerging from training on natural images. We find computer vision models, including large foundation models---regardless of training set, architecture, and supervision---struggle when multiple shortcuts are present. Even methods explicitly designed to combat shortcuts struggle in a Whac-A-Mole dilemma. To tackle this challenge, we propose Last Layer Ensemble, a simple-yet-effective method to mitigate multiple shortcuts without Whac-A-Mole behavior. Our results surface multi-shortcut mitigation as an overlooked challenge critical to advancing the reliability of vision systems. The datasets and code are released: \url{https://github.com/facebookresearch/Whac-A-Mole}.
\end{abstract}

\begin{figure*}
\centering
\begin{subfigure}[b]{.475\textwidth}
  \centering
  \includegraphics[width=\linewidth]{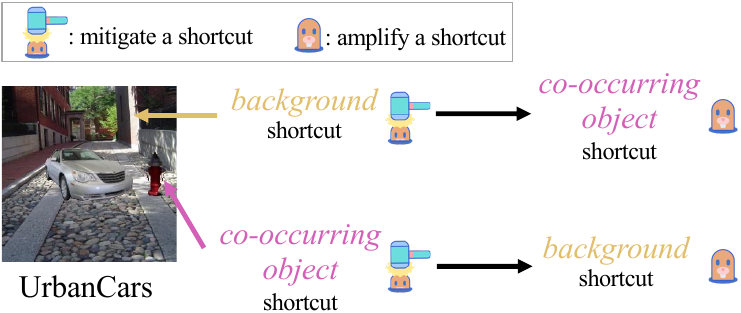}
  \caption{We construct UrbanCars, a new dataset with multiple shortcuts, facilitating the study of multi-shortcut learning under the \emph{controlled setting}.}
  \label{fig:teaser_urbancars}
\end{subfigure}%
\hfill
\begin{subfigure}[b]{.475\textwidth}
  \centering
  \includegraphics[width=\linewidth]{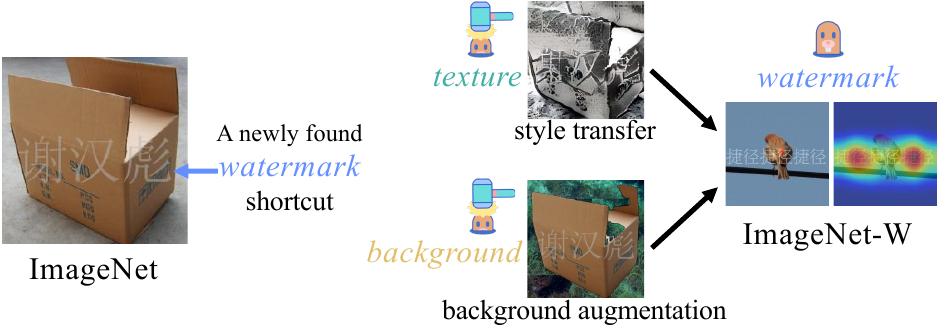}
  \caption{We discover the new watermark shortcut emerged from a \emph{natural image} dataset---ImageNet, and create ImageNet-W test set for ImageNet.}
  \label{fig:teaser_in_w}
\end{subfigure}
\caption{Our benchmark results on both datasets reveal the overlooked Whac-A-Mole dilemma in shortcut mitigation, \ie, mitigating one shortcut \molehammer{} amplifies the reliance on other shortcuts \molenohammer.}
\vspace{-6mm}
\end{figure*}

\section{Introduction}
Machine learning often achieves good average performance by exploiting unintended cues in the data~\cite{geirhos2020NatMachIntellShortcut}. For instance, when backgrounds are spuriously correlated with objects, image classifiers learn background as a rule for object recognition~\cite{xiao2021Int.Conf.Learn.Represent.Noise}.
This phenomenon---called ``shortcut learning''---at best suggests average metrics overstate model performance and at worst renders predictions unreliable as models are prone to costly mistakes on out-of-distribution (OOD) data where the shortcut is absent.
For example, COVID diagnosis models degraded significantly when spurious visual cues (\eg, hospital tags) were removed~\cite{degrave2021NatMachIntellAI}.

Most existing works design and evaluate methods
under the tenuous assumption that a \emph{single shortcut} is present in the data
~\cite{sagawa2020Int.Conf.Learn.Represent.Distributionally,nam2020Adv.NeuralInf.Process.Syst.Learning,he2021PatternRecognitionNonI}.
For instance, Waterbirds~\cite{sagawa2020Int.Conf.Learn.Represent.Distributionally}, the most widely-used dataset, only benchmarks the mitigation of the background shortcut
~\cite{liu2021Int.Conf.Mach.Learn.Just,creager2021Int.Conf.Mach.Learn.Environment,bao2022Int.Conf.Mach.Learn.Learning}.
While this is a useful simplified setting,
real-world images contain multiple visual cues;
models learn multiple shortcuts.
From ImageNet~\cite{singla2022Int.Conf.Learn.Represent.Salient,deng2009IEEEConf.Comput.Vis.PatternRecognit.CVPRImageNet} to facial attribute classification~\cite{lang2021IEEECVFInt.Conf.Comput.Vis.ICCVExplaining} and COVID-19 chest radiographs~\cite{degrave2021NatMachIntellAI},
multiple shortcuts are pervasive.
Whether existing methods can overcome multiple shortcuts or struggle in a \emph{Whac-A-Mole} game---where mitigating one shortcut amplifies others---remains a critical open question.

We directly address this limitation by proposing two datasets to study \emph{multi-shortcut} learning:
\textbf{UrbanCars}
and \textbf{ImageNet-W}.
In UrbanCars (\cref{fig:teaser_urbancars}), we precisely inject two spurious cues---background and co-occurring object.
UrbanCars allows us to conduct controlled experiments probing multi-shortcut learning in standard training as well as shortcut mitigation methods, including those requiring shortcut labels.
In ImageNet-W (IN-W) (\cref{fig:teaser_in_w}), we surface a new \emph{watermark} shortcut in the popular ImageNet dataset (IN-1k).
By adding a transparent watermark to IN-1k validation set images, ImageNet-W, as a new test set, reveals vision models ranging from ResNet-50~\cite{he2016IEEEConf.Comput.Vis.PatternRecognit.CVPRDeep} to large foundation models~\cite{bommasani2022Opportunities} \emph{universally rely on watermark as a spurious cue} for the ``carton'' class (\cf cardboard box in \cref{fig:teaser_in_w}).
When a watermark is added, ImageNet top-1 accuracy drops by 10.7\% on average across models. Some, such as ResNet-50, suffer a catastrophic 26.7\% drop (from 76.1\% on IN-1k to 49.4\% on IN-W) (\cref{subsec:imagenet_watermark})).
Along with texture ~\cite{geirhos2019Int.Conf.Learn.Represent.ImageNettrained,hendrycks2021IEEECVFInt.Conf.Comput.Vis.ICCVMany} and background~\cite{xiao2021Int.Conf.Learn.Represent.Noise} benchmarks, ImageNet-W allows us to study \emph{multiple shortcuts} emerging in natural images.

We find that across a range of supervised/self-supervised methods, network architectures, foundation models, and shortcut mitigation methods, vision models
struggle when multiple shortcuts are present.
Benchmarks on UrbanCars and multiple shortcuts in ImageNet (including ImageNet-W) reveal
an overlooked challenge in the shortcut learning problem: \textit{multi-shortcut mitigation resembles a Whac-A-Mole game, i.e., mitigating one shortcut amplifies reliance on others}.
Even methods specifically designed to combat shortcuts decrease reliance on one shortcut at the expense of amplifying others (\cref{sec:experiment}).
To tackle this open challenge, we propose Last Layer Ensemble (LLE) as the first endeavor to mitigate multiple shortcuts jointly without Whac-A-Mole behavior.
LLE uses data augmentation based on only the knowledge of the shortcut type without using shortcut labels---making it scalable to large-scale datasets.

To summarize, our contributions are (1) We create UrbanCars, a dataset with precisely injected spurious cues, to better benchmark multi-shortcut mitigation.
(2) We curate ImageNet-W---a new out-of-distribution (OOD) variant of ImageNet benchmarking a pervasive watermark shortcut we discovered---
to form a more comprehensive multi-shortcut evaluation suite for ImageNet.
(3) Through extensive benchmarks on UrbanCars and ImageNet shortcuts (including ImageNet-W), we uncover that mitigating multiple shortcuts is an overlooked and universal challenge,
resembling a Whac-A-Mole game, \ie, mitigating one shortcut amplifies reliance on others.
(4) Finally, we propose Last Layer Ensemble as the first endeavor for multi-shortcut mitigation without the Whac-A-Mole behavior.
We hope our contributions advance research into the overlooked challenge of mitigating multiple shortcuts.

\section{New Datasets for Multi-Shortcut Mitigation}

While most previous datasets~\cite{sagawa2020Int.Conf.Learn.Represent.Distributionally,arjovsky2020Invariant,nam2020Adv.NeuralInf.Process.Syst.Learning,liu2015IEEEInt.Conf.Comput.Vis.ICCVDeep} are based on the oversimplified single-shortcut setting, we introduce the UrbanCars dataset (\cref{subsec:urbancars}) and the ImageNet-Watermark dataset (\cref{subsec:imagenet_watermark}) to benchmark multi-shortcut mitigation.

\begin{figure}[!b]
  \vspace{-6mm}
  \centering
  \includegraphics[width=\linewidth]{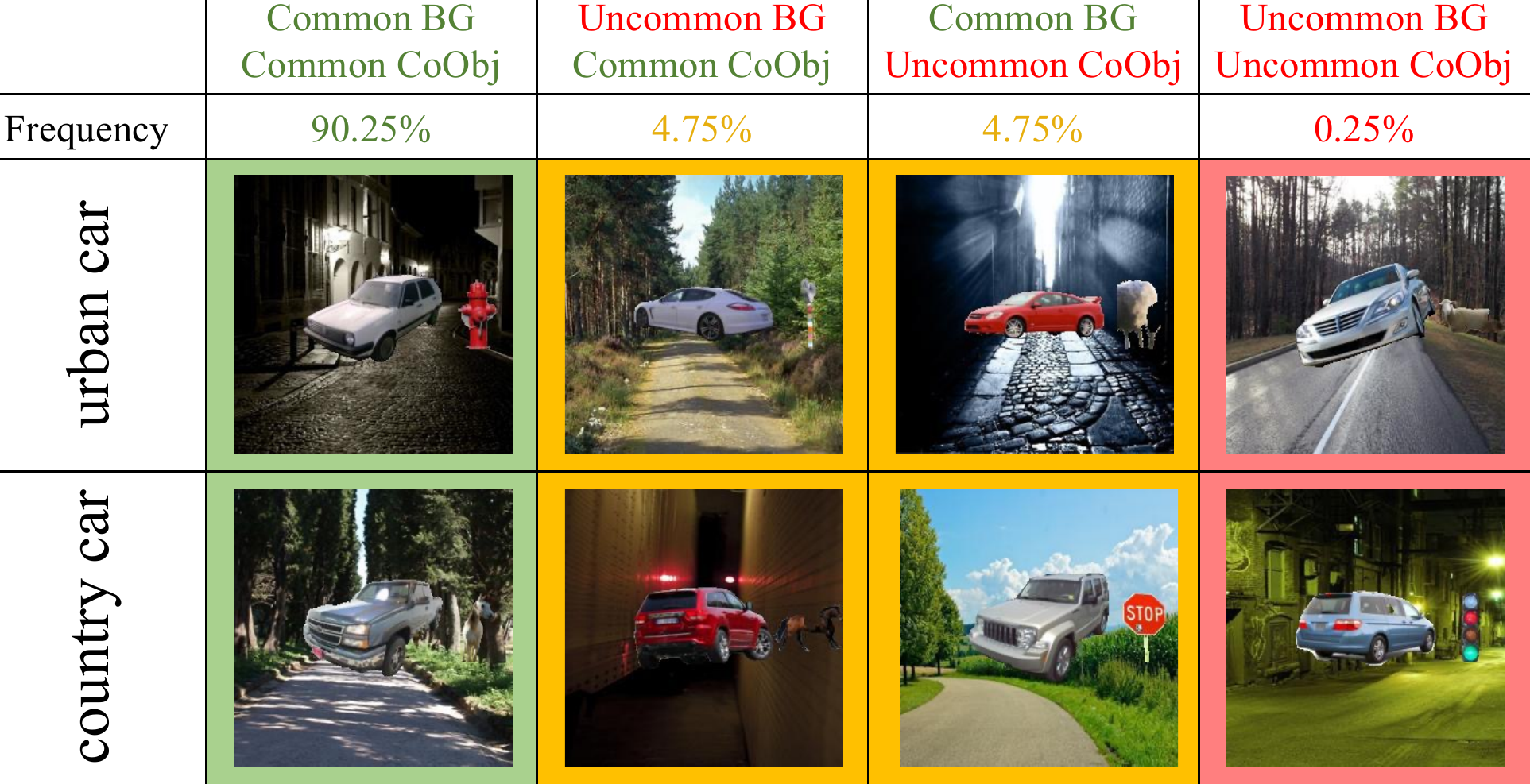}
  \caption{Unbalanced groups in UrbanCars's training set based on two shortcuts: \textit{background} (BG) and \textit{co-occurring object} (CoObj).}
   \label{fig:urbancars_details}
\end{figure}

\subsection{UrbanCars Dataset}
\label{subsec:urbancars}

\noindent \textbf{Overview} \quad We construct the UrbanCars dataset with multiple shortcuts: \textit{background} (BG) and \textit{co-occurring object} (CoObj). As shown in \cref{fig:urbancars_details}, each image in UrbanCars has a car at the center on a natural scene background with a co-occurring object on the right. The task is to classify the car's body type (\ie, target) by overcoming two shortcuts in the training set, which correlate with the target label.

Formally, we denote the dataset as a set of $N$ tuples, $\{ (x_i, y_i, b_i, c_i) \}_{i=1}^N$, where each image $x_i$ is annotated with three labels: target label $y_i$ for the car body type, \textit{background} label $b_i$, and \textit{co-occurring object} label $c_i$. We use a shared label space for all three labels with two classes: \texttt{urban} and \texttt{country}, \ie, $y_i, b_i, c_i \in \{ \texttt{urban}, \texttt{country} \}$. Based on the combination of three labels, the dataset is partitioned into $2^3=8$ groups, \ie, $\{\texttt{urban}, \texttt{country}\}$ car on the $\{\texttt{urban}, \texttt{country}\}$ BG with the $\{\texttt{urban}, \texttt{country}\}$ CoObj. We introduce the data distribution and construction below and include details in \cref{appx:subsec:urbancars_details}.

\noindent \textbf{Data Distribution} \quad The training set of UrbanCars has two spurious correlations of BG and CoObj shortcuts, whose strengths are quantified by $P(\mathbf{b} = \mathbf{y} \mid \mathbf{y})$ and $P(\mathbf{c} = \mathbf{y} \mid \mathbf{y})$, respectively. That is, the ratio of common BG (or CoObj) given a target class. We set both to 0.95 by following the correlation strength in \cite{sagawa2020Int.Conf.Learn.Represent.Distributionally}. We assume that two shortcuts are independently correlated with the target, \ie, $P(\mathbf{b}, \mathbf{c} \mid \mathbf{y}) = P(\mathbf{b} \mid \mathbf{y})P(\mathbf{c} \mid \mathbf{y})$.
As shown in \cref{fig:urbancars_details}, most urban car images have the urban background (\eg, alley) and urban co-occurring object (\eg, fire plug), and vice versa for country car images. The frequency of each group in the training set is in \cref{fig:urbancars_details}. The validation and testing sets are balanced without spurious correlations, \ie, ratios are 0.5.

\noindent \textbf{Data Construction} \quad The UrbanCars dataset is created from several source datasets.
The car objects and labels are from Stanford Cars~\cite{krause2013IEEEInt.Conf.Comput.Vis.Workshop3D}, where the urban cars are formed by classes such as sedan and hatchback. The country cars are from classes such as truck and van.
The backgrounds are from Places~\cite{zhou2018IEEETrans.PatternAnal.Mach.Intell.Places}. We use classes such as alley and crosswalk to form the urban background. The country background images are from classes such as forest road and field road.
Regarding co-occurring objects, we use LVIS~\cite{gupta2019IEEECVFConf.Comput.Vis.PatternRecognit.CVPRLVIS} to obtain the urban ones (\eg, fireplug and stop sign), and country ones (\eg, cow and horse). After obtaining the source images, we paste the car and co-occurring object onto the background.

\noindent \textbf{UrbanCars Metrics} \quad We first report the \textit{In Distribution Accuracy} (\textbf{I.D. Acc}) on UrbanCars. It computes the weighted average over accuracy per group, where weights are proportional to the training set's correlation strength (\ie, frequency in \cref{fig:urbancars_details}) by following ``average accuracy''~\cite{sagawa2020Int.Conf.Learn.Represent.Distributionally} to measure the performance when no group shift happens.

To measure robustness against the group shift, previous single-shortcut benchmarks ~\cite{creager2021Int.Conf.Mach.Learn.Environment,liu2021Int.Conf.Mach.Learn.Just,sagawa2020Int.Conf.Learn.Represent.Distributionally} use worst-group accuracy~\cite{sagawa2020Int.Conf.Learn.Represent.Distributionally}, \ie, the lowest accuracy among all groups. However, this metric does not capture multi-shortcut mitigation well since it only focuses on groups where both shortcut categories are uncommon (\cf the last column in \cref{fig:urbancars_details}).

To address this shortcoming, we introduce three new metrics: \textbf{BG Gap}, \textbf{CoObj Gap}, and \textbf{BG+CoObj Gap}. BG Gap is the accuracy drop from I.D. Acc to accuracy in groups where BG is uncommon but CoObj is common (\cf 1st yellow column in \cref{fig:urbancars_details}). Similarly, CoObj Gap computes the accuracy drop from I.D. Acc to groups where only CoObj is uncommon (\cf 2nd yellow column in \cref{fig:urbancars_details}). BG+CoObj Gap computes accuracy drop from I.D. Acc to groups where both BG and CoObj are uncommon (\cf red column in \cref{fig:urbancars_details}). The first two metrics measure the robustness against the group shift for each shortcut, and the last metric evaluates the model's robustness when both shortcuts are absent.

\begin{table*}[t]
\centering
\arrayrulecolor{lightgray}
\begin{adjustbox}{width=\linewidth}
\begin{tabular}{@{}lll|cc|cc|cc@{}}
             &              &                  & \multicolumn{2}{c|}{ \includegraphics[width=1in]{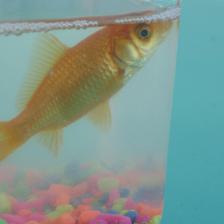} \includegraphics[width=1in]{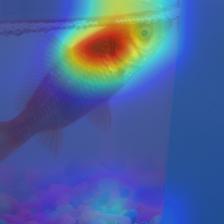} }            & \multicolumn{2}{c|}{ \includegraphics[width=1in]{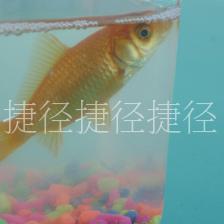} \includegraphics[width=1in]{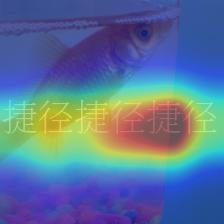} }                &  \multicolumn{2}{c}{\includegraphics[width=1in]{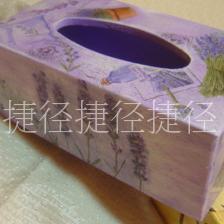} \includegraphics[width=1in]{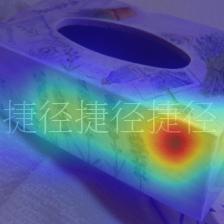}}                \\
             &              &                  & \multicolumn{2}{c|}{Prediction: \textcolor{ForestGreen}{goldfish} } & \multicolumn{2}{c|}{w/ Watermark: \textcolor{red}{carton}} & \multicolumn{2}{c}{w/ Watermark: \textcolor{red}{pencil sharpener} $\rightarrow$  \textcolor{ForestGreen}{carton}}   \\ \toprule \toprule
method  & architecture & (pre)training data                  & IN-1k Acc $\uparrow$ & $P(\hat{y} = \text{carton})$ (\%) & IN-W Gap $\uparrow$ & $\Delta P(\hat{y} = \text{carton})$ (\%) $\downarrow$ & Carton Gap $\downarrow$ & $\Delta P(\hat{y} = \text{carton} \mid y = \text{carton})$ (\%) $\downarrow$ \\ \midrule
Supervised     & ResNet-50~\cite{he2016IEEEConf.Comput.Vis.PatternRecognit.CVPRDeep}    & IN-1k~\cite{deng2009IEEEConf.Comput.Vis.PatternRecognit.CVPRImageNet}                       & 76.1       & 0.07           & -26.7 & +7.56  & +40 &  +42.46    \\
MoCov3~\cite{chen2021IEEECVFInt.Conf.Comput.Vis.ICCVEmpirical} (LP)     & ResNet-50    & IN-1k                       &  74.6  & 0.08                & -20.7 & +2.94  & +44 &  +44.37    \\
Style Transfer~\cite{geirhos2019Int.Conf.Learn.Represent.ImageNettrained}  & ResNet-50    & SIN~\cite{geirhos2019Int.Conf.Learn.Represent.ImageNettrained}                       & 60.1   &   0.10             & -17.3 & +4.91  & +52 &   +50.06    \\
Mixup~\cite{zhang2018Int.Conf.Learn.Represent.mixup}     & ResNet-50    & IN-1k                       &  76.1  & 0.07                & -18.6 & +3.43  & +38 &  +39.78    \\
CutMix~\cite{yun2019IEEECVFInt.Conf.Comput.Vis.ICCVCutMix}     & ResNet-50    & IN-1k                       &  78.5  & 0.09                & -14.8 & +1.92  & +22 &  +29.61    \\
Cutout~\cite{devries2017Improved,zhong2020AAAIConf.Artif.Intell.Random}     & ResNet-50    & IN-1k                       &  77.0  & 0.08                & -18.0 & +2.93  & +32 &  +38.06    \\
AugMix~\cite{hendrycks2020Int.Conf.Learn.Represent.AugMixa}     & ResNet-50    & IN-1k                       &  77.5  & 0.09                & -16.8 & +2.61  & +36 &  +34.44    \\
\midrule
Supervised     & RG-32gf      & IN-1k                       & 80.8  &   0.09                    & -14.1  &  +3.74     & +32 &     +33.43       \\
SEER~\cite{goyal2022Vision} (FT) & RG-32gf~\cite{radosavovic2020IEEECVFConf.Comput.Vis.PatternRecognit.CVPRDesigning}      & IG-1B~\cite{goyal2022Vision}  & 83.3 &    0.09                    & -6.5  &  +0.56       & +18   &  +24.26        \\
\midrule
Supervised       & ViT-B/32~\cite{dosovitskiy2021Int.Conf.Learn.Represent.Image}     & IN-1k      & 75.9     & 0.09      & -8.7         & +1.20              & +34        & +34.31                      \\
Uniform Soup~\cite{wortsman2022Int.Conf.Mach.Learn.Model} (FT) & ViT-B/32        & WIT~\cite{radford2021Int.Conf.Mach.Learn.Learning}   & 79.9       &   0.09      & -7.9  &  +0.32      & +24  &   +23.87        \\ Greedy Soup~\cite{wortsman2022Int.Conf.Mach.Learn.Model} (FT) & ViT-B/32        & WIT  & 81.0       &   0.09      & -6.5  &  +0.35      & +16  &   +23.87        \\
\midrule
Supervised    & ViT-L/16        & IN-1k   & 79.6    &   0.08      & -6.2 & +0.82 & +34 & +32.57 \\
CLIP~\cite{radford2021Int.Conf.Mach.Learn.Learning} (zero-shot)   & ViT-L/14        & WIT                     & 76.5    &   0.06                  & -4.4   &  \textbf{+0.01}     & +12   & \textbf{+1.75} \\
CLIP (zero-shot)    & ViT-L/14        & LAION-400M~\cite{schuhmann2021Adv.NeuralInf.Process.Syst.WorkshopLAION400M}                      & 72.7   &    0.05                  & -4.9   &   +0.03    & +12 & +13.76  \\ \midrule
MAE~\cite{he2022IEEECVFConf.Comput.Vis.PatternRecognit.CVPRMasked} (FT)    & ViT-H/14        & IN-1k                    & 86.9  &     0.08                  & -3.5 & +0.43 & +30     &  +29.59    \\
SWAG~\cite{singh2022IEEECVFConf.Comput.Vis.PatternRecognit.CVPRRevisiting} (LP)    & ViT-H/14        & IG-3.6B~\cite{singh2022IEEECVFConf.Comput.Vis.PatternRecognit.CVPRRevisiting}                    & 85.7    &   0.09                  & -4.9 & +0.19 & \textbf{+8} & +12.80 \\
SWAG (FT)    & ViT-H/14        & IG-3.6B                    & 88.5    &   0.09                  & \textbf{-3.1} & +0.35 & +18 & +20.25 \\
CLIP (zero-shot)    & ViT-H/14        & LAION-2B~\cite{schuhmann2022LAION5B}                    & 77.9      &   0.06                & -3.6 & +0.03  & +16    &  +12.01       \\ \midrule
\midrule
\multicolumn{3}{c|}{average}                 & 78.6      &   0.08                & -10.7 & +1.74  & +26.7    &  +27.96
\\ \bottomrule
\end{tabular}
\end{adjustbox}
\caption{\textbf{Models rely on the watermark as a shortcut for the carton class.} LP and FT denote linear probing and fine-tuning on ImageNet-1k, respectively. Because models exhibit drops (\ie, IN-W Gap) and an increase in accuracy and predicted probability of the carton class from IN-1k to IN-W, we conclude that various vision models suffer from the watermark shortcut (more results in \cref{appx:subset:in_w_results_more_methods,appx:subsec:in_w_on_in_v2}).}
\label{tab:watermark}
\vspace{-6mm}
\end{table*}

\begin{figure}[t]
  \centering
  \includegraphics[width=\linewidth]{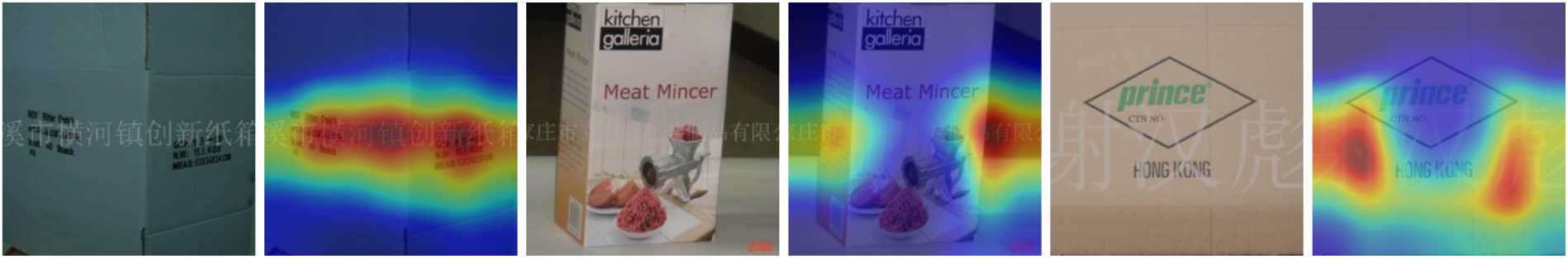}
   \caption{Many carton class images in the ImageNet training set contain the watermark. Saliency maps~\cite{selvaraju2017IEEEInt.Conf.Comput.Vis.ICCVGradCAM} of ResNet-50~\cite{he2016IEEEConf.Comput.Vis.PatternRecognit.CVPRDeep} show that the watermark serves as the shortcut for the carton class.}
   \label{fig:watermark_IN_1K_carton_train}
   \vspace{-6mm}
\end{figure}

\subsection{ImageNet-Watermark (ImageNet-W)}
\label{subsec:imagenet_watermark}

In addition to the precisely controlled spurious correlations in UrbanCars, we study naturally occurring shortcuts in the most popular computer vision benchmark: ImageNet~\cite{deng2009IEEEConf.Comput.Vis.PatternRecognit.CVPRImageNet}.
While ImageNet lacks shortcut labels, we can evaluate models' reliance on texture~\cite{geirhos2019Int.Conf.Learn.Represent.ImageNettrained} and background~\cite{xiao2021Int.Conf.Learn.Represent.Noise} shortcuts.
We additionally discovered a pervasive watermark shortcut and
contribute ImageNet-Watermark (ImageNet-W or IN-W), an evaluation set to expose models' watermark shortcut reliance. Along with texture and background, this forms a comprehensive suite
to evaluate reliance on the multiple naturally occurring shortcuts in ImageNet.

\noindent \textbf{Watermark Shortcut in ImageNet} \quad In the training set of the \textit{carton} class, many images contain a watermark at the center written in Chinese characters and ImageNet-trained ResNet-50~\cite{he2016IEEEConf.Comput.Vis.PatternRecognit.CVPRDeep} focuses on the watermark region to predict the carton class (\cref{fig:watermark_IN_1K_carton_train}). Since the watermark reads carton factory names or contact person's names of a carton factory, we conjecture that this watermark shortcut originates from the real-world spurious correlation of web images. In the validation set, none of the carton class images contain the watermark, so ResNet-50 underperforms on the carton class (48\%) relative to overall accuracy (76\%) across 1k classes.

\begin{CJK*}{UTF8}{gbsn}
\noindent \textbf{Data Construction} \quad To test the robustness against the watermark shortcut, we create ImageNet-Watermark (ImageNet-W or IN-W) dataset, a new out-of-distribution evaluation set of ImageNet. As shown in \cref{tab:watermark}, we overlay a transparent watermark written in ``捷径捷径捷径'' at the center of all images from ImageNet validation set to mimic the watermark pattern in IN-1k, where ``捷径'' means ``shortcut'' in Chinese.
We do this because we find that models use the watermark even when the content is not identical to the watermark in the training set of carton images, suggesting that it is watermark's presence rather than its content that serves as the shortcut. We evaluate watermark in other contents and languages in \cref{appx:subsec:in_w_details}.
\end{CJK*}

\begin{figure}[t]
  \centering
  \includegraphics[width=\linewidth]{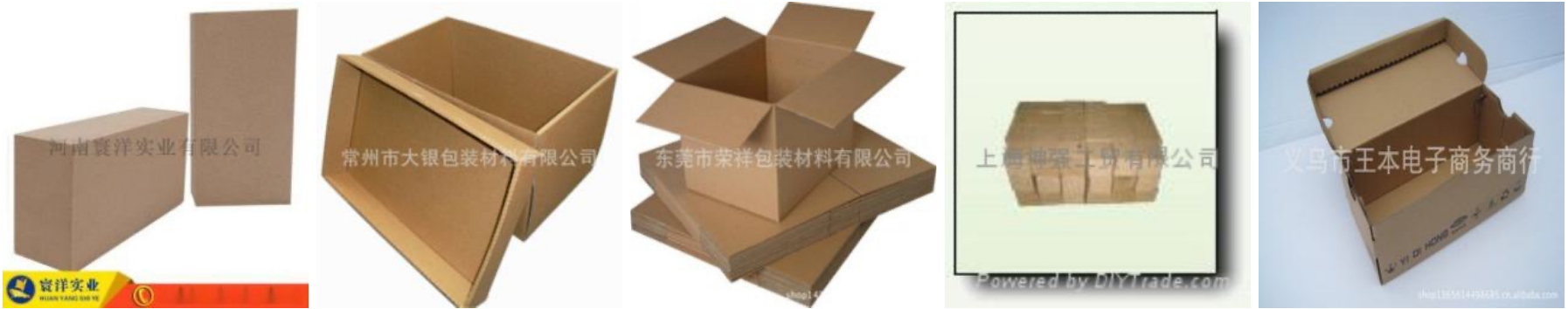}
   \caption{Carton images from LAION~\cite{schuhmann2021Adv.NeuralInf.Process.Syst.WorkshopLAION400M,schuhmann2022LAION5B}, a large-scale dataset with 400 million to 2 billion images used in CLIP~\cite{radford2021Int.Conf.Mach.Learn.Learning} pretraining, also contain watermarks, enabling CLIP's reliance on the watermark shortcut in zero-shot transfer to ImageNet and ImageNet-W.}
   \label{fig:laion_carton}
   \vspace{-6mm}
\end{figure}

\noindent \textbf{ImageNet-W Metrics} \quad We mainly use two metrics to measure watermark shortcut reliance:
(1) \textbf{IN-W~Gap} is the accuracy on IN-W minus the accuracy on IN-1k validation set. A smaller accuracy drop indicates less reliance on the watermark shortcut across all 1000 classes. (2) \textbf{Carton Gap} is the carton class accuracy increase from IN-1k to IN-W. A smaller Carton Gap indicates less reliance on the watermark shortcut for predicting the carton class.

To demonstrate that the watermark shortcut is used for predicting carton, we use the following in \cref{tab:watermark}: (1) $P(\hat{y} = \text{carton})$, the predicted probability of carton on all IN-1k validation set images, (2) $\Delta P(\hat{y} = \text{carton})$, the predicted probability increase from IN-1k to IN-W of all 1k classes, and (3) $\Delta P(\hat{y} = \text{carton} \mid y = \text{carton})$, the predicted probability increase from IN-1k to IN-W of the carton class.

\noindent \textbf{Ubiquitous reliance on the watermark shortcut} \quad To study reliance on the watermark shortcut, we use ImageNet-W to benchmark a broad range of State-of-The-Art (SoTA) vision models, including standard supervised training, using
different architectures~\cite{he2016IEEEConf.Comput.Vis.PatternRecognit.CVPRDeep,radosavovic2020IEEECVFConf.Comput.Vis.PatternRecognit.CVPRDesigning,dosovitskiy2021Int.Conf.Learn.Represent.Image},  augmentations and regularizations~\cite{geirhos2019Int.Conf.Learn.Represent.ImageNettrained,zhang2018Int.Conf.Learn.Represent.mixup,yun2019IEEECVFInt.Conf.Comput.Vis.ICCVCutMix,hendrycks2020Int.Conf.Learn.Represent.AugMixa}. We also benchmark foundation models~\cite{bommasani2022Opportunities} pretrained on larger datasets~\cite{singh2022IEEECVFConf.Comput.Vis.PatternRecognit.CVPRRevisiting,schuhmann2022LAION5B,goyal2022Vision,schuhmann2021Adv.NeuralInf.Process.Syst.WorkshopLAION400M,radford2021Int.Conf.Mach.Learn.Learning} with different pretraining supervision and transfer learning techniques~\cite{chen2021IEEECVFInt.Conf.Comput.Vis.ICCVEmpirical,he2022IEEECVFConf.Comput.Vis.PatternRecognit.CVPRMasked,goyal2022Vision,singh2022IEEECVFConf.Comput.Vis.PatternRecognit.CVPRRevisiting,radford2021Int.Conf.Mach.Learn.Learning,wortsman2022Int.Conf.Mach.Learn.Model}.
In \cref{tab:watermark}, we find a considerable IN-W Gap of up to -26.7 and -10.7 on average and a Carton Gap of up to +52 and +26.7 on average.
While all models exhibit uniform ($1/1000=0.1\%$) predicted probabilities for carton class ($P(\hat{y} = \text{carton})$) on IN-1k, we observe a considerable increase in the predicted probability of carton on IN-W ($\Delta P(\hat{y} = \text{carton})$) and a significant predicted probability increase in carton class images ($\Delta P(\hat{y} = \text{carton} \mid y = \text{carton})$). Although compared to supervised ResNet-50, some models with larger architectures or extra training data can decrease reliance on the watermark shortcut, none of them fully close the performance gaps.
Interestingly, CLIP with zero-shot transfer still suffers from the watermark shortcut with +12 to +16 Carton Gap, which could be explained by many carton images in the pretraining data (\eg, LAION) also containing watermarks (\cf \cref{fig:laion_carton}). To the best of our knowledge, this is the first real-world example of \textbf{the existence of shortcut in billion-scale datasets for foundation model pretraining}, which also confirms findings that data quality, not quantity~\cite{fang2022Int.Conf.Mach.Learn.Data,nguyen2022Adv.NeuralInf.Process.Syst.Quality}, matters most to CLIP's robustness.

\noindent \textbf{Multi-Shortcut Mitigation Metrics on ImageNet} \quad To measure the mitigation of multiple shortcuts, we evaluate models on multiple OOD variants of ImageNet. In this work, we study three shortcuts on ImageNet---background, texture, and watermark. The background shortcut is evaluated on ImageNet-9 (IN-9)~\cite{xiao2021Int.Conf.Learn.Represent.Noise}, and we use \textbf{IN-9~Gap} (\ie, BG-Gap in \cite{xiao2021Int.Conf.Learn.Represent.Noise}) as the evaluation metric, which is the accuracy drop from Mixed-Same to Mixed-Rand in IN-9, where a lower accuracy drop implies less background shortcut reliance. The texture shortcut is evaluated on Stylized ImageNet (SIN)~\cite{geirhos2019Int.Conf.Learn.Represent.ImageNettrained} and ImageNet-R (IN-R)~\cite{hendrycks2021IEEECVFInt.Conf.Comput.Vis.ICCVMany}, where we use \textbf{SIN Gap}, top-1 accuracy drop from IN-1k to SIN, and \textbf{IN-R Gap}, the top-1 accuracy drop from IN-200 (\ie, a subset of IN-1k with 200 classes used in IN-R) to IN-R.

\section{Benchmark Methods and Settings}
\label{subsec:benchmark_and_comparison_methods}

On all datasets, we first evaluate standard training that minimizes the empirical risk on the training set (\ie, \textbf{ERM}~\cite{vapnik1999Nature}) using ResNet-50~\cite{he2016IEEEConf.Comput.Vis.PatternRecognit.CVPRDeep} as the network architecture, which serves as the baseline. On ImageNet, we additionally show ERM's results with other architectures, pretraining datasets, and supervision.

In addition to ERM, we comprehensively evaluate shortcut mitigation methods across four categories based on the level of shortcut information required (\cref{tab:benchmark_methods_summary}).

\begin{table}[b]
\vspace{-4mm}
\centering
\begin{adjustbox}{width=\linewidth}
\begin{tabular}{@{}p{0.15\linewidth}|p{0.3\linewidth}|p{0.4\linewidth}|p{0.475\linewidth}@{}}
\toprule
\textbf{Category} & \textbf{Summary}                                & \textbf{Shortcut Information} & \textbf{Methods}                                                   \\ \midrule
1        & Standard Augmentation and Regularization & None                    & Mixup~\cite{zhang2018Int.Conf.Learn.Represent.mixup}, Cutout~\cite{devries2017Improved,zhong2020AAAIConf.Artif.Intell.Random}, CutMix~\cite{yun2019IEEECVFInt.Conf.Comput.Vis.ICCVCutMix}, AugMix~\cite{hendrycks2020Int.Conf.Learn.Represent.AugMixa}, SD~\cite{pezeshki2021Adv.NeuralInf.Process.Syst.Gradient}                         \\ \midrule
2        & Targeted Augmentation for Mitigating Shortcuts         & Types of shortcuts (w/o shortcut labels)        & CF+F Aug~\cite{chang2021IEEECVFConf.Comput.Vis.PatternRecognit.CVPRRobust}, Style Transfer (TXT Aug)~\cite{geirhos2019Int.Conf.Learn.Represent.ImageNettrained}, BG Aug~\cite{xiao2021Int.Conf.Learn.Represent.Noise,ryali2021Characterizing}, WMK Aug \\ \midrule
3        & Using Shortcut Labels                   & Image-level ground-truth shortcut label      & gDRO~\cite{sagawa2020Int.Conf.Learn.Represent.Distributionally}, DI~\cite{wang2020IEEECVFConf.Comput.Vis.PatternRecognit.CVPRFairness}, SUBG~\cite{idrissi2022Conf.CausalLearn.Reason.Simple}, DFR~\cite{kirichenko2022Last}                                       \\ \midrule
4        & Inferring Pseudo Shortcut Labels        & Image-level pseudo shortcut label           & LfF~\cite{nam2020Adv.NeuralInf.Process.Syst.Learning}, JTT~\cite{liu2021Int.Conf.Mach.Learn.Just}, EIIL~\cite{creager2021Int.Conf.Mach.Learn.Environment}, DebiAN~\cite{li2022Eur.Conf.Comput.Vis.ECCVDiscovera}                                    \\ \bottomrule
\end{tabular}
\end{adjustbox}
\caption{Existing methods for multi-shortcut mitigation benchmark.}
\label{tab:benchmark_methods_summary}
\end{table}

\noindent \textbf{Category 1: Standard Augmentation and Regularization} \quad Methods in this category use general data augmentation or regularization without prior knowledge of the shortcut, which are commonly used to improve accuracy on IN-1k, \eg, new training recipes~\cite{vryniotis2021PyTorchBlogHow,wightman2021ResNet}.
Some works~\cite{chang2021IEEECVFConf.Comput.Vis.PatternRecognit.CVPRRobust,pinto2022Adv.NeuralInf.Process.Syst.RegMixup} show that they can also improve OOD robustness.

\noindent \textbf{Category 2: Targeted Augmentation for Mitigating Shortcuts} \; Other works use data augmentation that modifies shortcut cues. We evaluate CF+F Aug~\cite{chang2021IEEECVFConf.Comput.Vis.PatternRecognit.CVPRRobust} on UrbanCars. On ImageNet, we benchmark texture augmentation (TXT Aug) via style transfer~\cite{geirhos2019Int.Conf.Learn.Represent.ImageNettrained} and background augmentation (BG Aug)~\cite{xiao2021Int.Conf.Learn.Represent.Noise,ryali2021Characterizing}. To counter the watermark shortcut, we design watermark augmentation (WTM Aug) that randomly overlays the watermark onto images (\cf \cref{appx:subsec:watermark_aug}).

\noindent \textbf{Category 3: Using Shortcut Labels} \quad In this category, methods use shortcut labels for mitigation, which are generally used to reweight~\cite{sagawa2020Int.Conf.Learn.Represent.Distributionally} or resample training data~\cite{sagawa2020Int.Conf.Learn.Represent.Distributionally,idrissi2022Conf.CausalLearn.Reason.Simple,kirichenko2022Last}. We only benchmark methods in this category on UrbanCars since ImageNet does not have shortcut labels.

\noindent \textbf{Category 4: Inferring Pseudo Shortcut Labels} \quad Following the ideas of methods using shortcut labels, one line of works~\cite{liu2021Int.Conf.Mach.Learn.Just,creager2021Int.Conf.Mach.Learn.Environment,nam2020Adv.NeuralInf.Process.Syst.Learning,li2022Eur.Conf.Comput.Vis.ECCVDiscovera} estimates the pseudo shortcut labels when ground-truth labels are unavailable.

\noindent \textbf{Benchmark Settings} \quad We introduce the experiment settings here (details in \cref{appx:subsec:detailed_exp_setting}).
On UrbanCars, we use worst-group accuracy~\cite{sagawa2020Int.Conf.Learn.Represent.Distributionally} on the validation set to select the early stopping epoch and report test set results.
All methods except DFR~\cite{kirichenko2022Last} use end-to-end training on UrbanCars.
On ImageNet, following the last layer re-training~\cite{kirichenko2022Last} setting, we only train the last classification layer upon a frozen feature extractor.
On both datasets, we use ResNet-50 as the network architecture. On ImageNet, we also benchmark self-supervised and foundation models.

\section{Our Approach}
\label{sec:our_approach_lle}

\noindent \textbf{Motivation} \quad Our multi-shortcut benchmark results (\cref{sec:experiment}) show that many existing methods suffer from the Whac-A-Mole problem, motivating us to design a method to mitigate multiple shortcuts simultaneously.

We focus on mitigating multiple \textit{known} shortcuts---the number and types of shortcuts are given, but shortcut labels are not. The absence of shortcut labels makes it scalable to large datasets (\eg, ImageNet). Although mitigating unknown numbers and types of shortcuts seems more desirable, not only do our empirical results show their under-performance, but also it is theoretically impossible to mitigate shortcuts without any inductive biases~\cite{lin2022Adv.NeuralInf.Process.Syst.ZINa}.

We follow methods that use data augmentation to modify the shortcut cues (\ie, category 2). Formally, given a set of $K$ shortcuts $\{ \mathrm{s}_i \}_{i=1}^K$ for mitigation, we create a set of augmentations $\mathrm{S}_\text{aug} =  \{ \mathcal{A}_i \}_{i=1}^K \cup \{ \mathcal{I} \}$, where the augmentation $\mathcal{A}_i$ (\eg, style transfer~\cite{geirhos2019Int.Conf.Learn.Represent.ImageNettrained}) modifies the visual cue of the shortcut $ \mathrm{s}_i$ (\eg, texture). $\mathcal{I}$ denotes the identity transformation, \ie, no augmentation applied.

Based on the augmentation set $\mathrm{S}_\text{aug}$, a straightforward way is to minimize the empirical risk~\cite{vapnik1999Nature} over all augmented and original images.
However, different augmentations can be incompatible, leading to suboptimal results. That is, augmentation $\mathcal{A}_i$ could be detrimental to mitigating a different shortcut $ \mathrm{s}_j$, where $i \neq j$. For example, mitigating the texture shortcut via style transfer~\cite{geirhos2019Int.Conf.Learn.Represent.ImageNettrained} augmentation unexpectedly amplifies the saliency of the watermark (\cref{fig:teaser_in_w}), leading to worse watermark mitigation results (\cref{tab:watermark}).

\begin{figure}[t]
  \centering
  \includegraphics[width=0.95\linewidth]{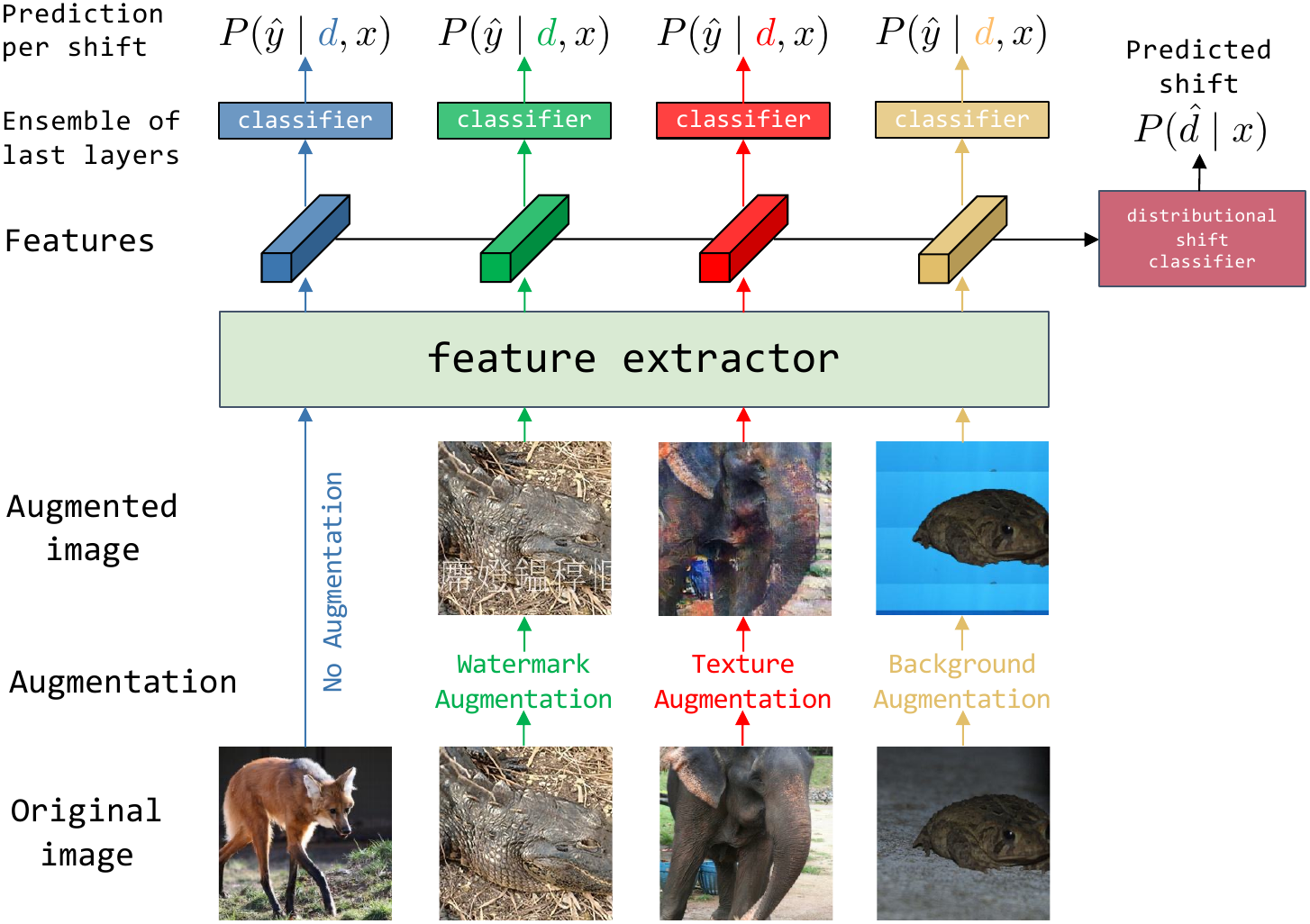}
  \caption{An overview of Last Layer Ensemble (LLE). LLE trains an ensemble of the last classification layers upon a feature extractor, where each last layer is trained with images in one augmentation type. The distributional shift classifier, supervised by the augmentation type, is trained to predict the distributional shift and dynamically aggregates the predictions per shift during testing.}
  \label{fig:method_last_layer_ensemble}
  \vspace{-6mm}
\end{figure}

\noindent \textbf{Last Layer Ensemble} \quad To address this issue, we propose Last Layer Ensemble (LLE), a new method for mitigating multiple shortcuts simultaneously (\cref{fig:method_last_layer_ensemble}). Since it is hard to use a single model to learn the invariance among incompatible augmentations, we instead train an ensemble~\cite{dietterich2000Mult.Classif.Syst.Ensemble} of classification layers (\ie, last layers) on top of a shared feature extractor so that each classification layer only trains on data from a single type of augmentation that simulates one type of distributional shift $d$. In this way, each last layer predicts the probability of the target $P(\hat{y} \mid d, x)$.

At the same time, we train a \textit{distributional shift classifier}, another classification layer on top of the feature extractor, to predict the type of augmentation that simulates the distributional shift, \ie, $P(\hat{d} \mid x)$.
During testing, LLE dynamically aggregates the logits from the ensemble of the last layers based on the predicted distributional shift. \Eg, when the testing image contains the texture shift, the \textit{distributional shift classifier} gives higher weights for the logits from the classifier trained with texture augmentation, alleviating the impact from other classification layers trained with incompatible augmentations. In addition, when the weights of the feature extractor are not frozen, we stop the gradient from the \textit{distributional shift classifier} to the feature extractor, preventing the feature extractor from learning the shortcut information. Compared to standard ensemble approaches~\cite{dietterich2000Mult.Classif.Syst.Ensemble} that train multiple full networks and add significant inference overhead, our method uses minimal additional training parameters and has better computational efficiency.

\begin{table}[t]
\centering
\begin{adjustbox}{width=\linewidth}
\begin{tabular}{@{}lclll@{}}
\toprule
                    & \multicolumn{1}{c|}{}                    & \multicolumn{3}{c}{shortcut reliance}               \\ \cmidrule(l){3-5}
                    & \multicolumn{1}{c|}{I.D. Acc} & BG Gap $\uparrow$      & CoObj Gap $\uparrow$  & BG+CoObj Gap $\uparrow$   \\ \midrule
\textcolor{gray}{ERM}        & \textcolor{gray}{97.6}        & \textcolor{gray}{-15.3}         & \textcolor{gray}{-11.2}         & \textcolor{gray}{-69.2}        \\ \midrule
Mixup               & 98.3                                   & -12.6         & -9.3          & -61.8         \\
CutMix              & 96.6                                   & -45.0 (\textcolor{red}{$\times$2.94} \molenohammer)        & -4.8          & -86.5         \\
Cutout              & 97.8                                   & -15.8 (\textcolor{red}{$\times$1.03} \molenohammer)        & -10.4         & -71.4         \\
AugMix              & 98.2                                   & -10.3         & -12.1 (\textcolor{red}{$\times 1.08$} \molenohammer)         & -70.2         \\
SD                  & 97.3                                   & -15.0         & -3.6          & -36.1         \\ \midrule
CF+F Aug            & 96.8                                   & -16.0 (\textcolor{red}{$\times$1.04} \molenohammer)         & \textbf{+0.4}         & -19.4         \\ \midrule
LfF                 & 97.2                                   & -11.6          & -18.4 (\textcolor{red}{$\times$1.64} \molenohammer)        & -63.2         \\
JTT (E=1)           & 95.9                                   & -8.1          & -13.3 (\textcolor{red}{$\times$1.18} \molenohammer)        & -40.1         \\
EIIL (E=1)          & 95.5                                   & -4.2          & -24.7 (\textcolor{red}{$\times$2.21} \molenohammer)        & -44.9         \\
JTT (E=2)           & 94.6                                   & -23.3  (\textcolor{red}{$\times$1.52} \molenohammer)       & -5.3          & -52.1         \\
EIIL (E=2)          & 95.5                                   & -21.5 (\textcolor{red}{$\times$1.40} \molenohammer)        & -6.8          & -49.6         \\
DebiAN              & 98.0                                   & -14.9         & -10.5         & -69.0         \\ \midrule \midrule
\textbf{LLE (ours)} & 96.7                                   & \textbf{-2.1} & -2.7 & \textbf{-5.9} \\ \bottomrule
\end{tabular}
\end{adjustbox}
\caption{\textbf{Many methods not using shortcut labels (category 1,2,4) amplify shortcut on UrbanCars.} \molenohammer: increased reliance on a shortcut relative to ERM. $\times2.94$: 2.94 times larger than ERM.
}
\label{tab:urbancars_unsupervised_results}
\vspace{-4mm}
\end{table}

\section{Experiments}
\label{sec:experiment}

Based on UrbanCars and ImageNet-W datasets, we show results on multi-shortcut mitigation.
We first study if standard supervised training (\ie, ERM) relies on multiple shortcuts (\cref{subsec:standard_training}).
Next, we show the multi-shortcut setting is significantly challenging: mitigating one shortcut increases reliance on other shortcuts compared to ERM.
We name this phenomenon \emph{Whac-A-Mole}, which is observed in many SoTA methods, including mitigation methods (\cref{subsec:mitigation_methods_results}) and self-supervised/foundation models (\cref{subsec:self_sup_foundation_results}).
Finally, we show that our Last Layer Ensemble method can reduce reliance across multiple shortcuts more effectively (\cref{subsec:lle_results}).

\subsection{Standard training relies on multiple shortcuts}
\label{subsec:standard_training}

On both datasets, we find that standard training (\ie, ERM~\cite{vapnik1999Nature}) relies on multiple shortcuts.
On UrbanCars, \cref{tab:urbancars_unsupervised_results} shows that ERM achieves near zero in-distributional error (97.6\% I.D. Acc.).
However, ERM's performance drops when group shift happens.
When the background shortcut is absent, ERM's performance drops by 15.3\% in BG Gap.
Similarly, the accuracy drops by 11.2\% in CoObj Gap when the CoObj shortcut is absent.
When neither shortcut is present, models suffer catastrophic drops of 69.2\% in BG+CoObj Gap.
On ImageNet, \cref{tab:imagenet_results} shows that ERM achieves good top-1 accuracy of 76.39\% on IN-1k.
However, it suffers considerable drops in accuracy when watermark, texture, or background cues are altered, \eg, 30\% Carton Gap for watermark, 56-69\% for texture, and 5.19\% for background, suggesting that standard training on natural images from ImageNet leads to reliance on multiple shortcuts.

\begin{table}[t]
\centering
\begin{adjustbox}{width=\linewidth}
\begin{tabular}{@{}lllllll@{}}
\toprule
                          & \multicolumn{1}{l|}{}      & \multicolumn{5}{c}{shortcut reliance}                                                                                                                                 \\ \cmidrule(l){3-7}
             & \multicolumn{1}{l|}{}      & \multicolumn{2}{c|}{Watermark (WTM)}                                     & \multicolumn{2}{c|}{Texture (TXT)}                                             & Background (BG)            \\
                           & \multicolumn{1}{l|}{IN-1k} & IN-W Gap $\uparrow$ & \multicolumn{1}{l|}{Carton Gap $\downarrow$} & SIN Gap  $\uparrow$           & \multicolumn{1}{l|}{IN-R Gap $\uparrow$} & IN-9 Gap   $\uparrow$ \\ \midrule
\textcolor{gray}{ERM}        & \textcolor{gray}{76.39}       & \textcolor{gray}{-25.40}     & \textcolor{gray}{+30}      & \textcolor{gray}{-69.43}         & \textcolor{gray}{-56.22}        & \textcolor{gray}{-5.19}                \\ \midrule
Mixup                           & 76.17                & -24.87               & +34 \footnotesize{(\textcolor{red}{$\times 1.13$} \molenohammer)}               & -68.18               & -55.79               & -5.60  \footnotesize{(\textcolor{red}{$\times 1.08$} \molenohammer)}             \\
CutMix                          & 75.90                & -25.78 \footnotesize{(\textcolor{red}{$\times 1.01$} \molenohammer)}             & +32 \footnotesize{(\textcolor{red}{$\times 1.06$} \molenohammer)}                 & -69.31               & -56.36               & -5.65 \footnotesize{(\textcolor{red}{$\times 1.09$} \molenohammer)}             \\
Cutout                      & 76.40                & -25.11               & +32 \footnotesize{(\textcolor{red}{$\times 1.06$} \molenohammer)}                 & -69.39               & -55.93               & -5.35   \footnotesize{(\textcolor{red}{$\times 1.03$} \molenohammer)}         \\
AugMix                      & 76.23                & -23.41               & +38  \footnotesize{(\textcolor{red}{$\times 1.26$} \molenohammer)}                & -68.51               & -54.91               & -5.85 \footnotesize{(\textcolor{red}{$\times 1.13$} \molenohammer)}              \\
SD                    & 76.39                & -26.03 \footnotesize{(\textcolor{red}{$\times 1.02$} \molenohammer)}               & +30                  & -69.42               & -56.36      & -5.33  \footnotesize{(\textcolor{red}{$\times 1.03$} \molenohammer)}              \\ \midrule
WTM Aug                 & 76.32                & \textbf{-5.78}       & +14                  & -69.31               & -56.22               & -5.34   \footnotesize{(\textcolor{red}{$\times 1.03$} \molenohammer)}             \\
TXT~\molehammer Aug                  & 75.94                & -25.93 \footnotesize{(\textcolor{red}{$\times 1.02$} \molenohammer)}       & +36  \footnotesize{(\textcolor{red}{$\times 1.20$} \molenohammer)}                & -63.99               & \textbf{-53.24}               & -5.66 \footnotesize{(\textcolor{red}{$\times 1.09$} \molenohammer)}               \\
BG~\molehammer Aug                & 76.03                & -25.01               & +36 \footnotesize{(\textcolor{red}{$\times 1.20$} \molenohammer)}                 & -68.41               & -54.51               & -4.67                \\ \midrule
LfF                        & 76.35                & -26.19  \footnotesize{(\textcolor{red}{$\times 1.03$} \molenohammer)}              & +36   \footnotesize{(\textcolor{red}{$\times 1.20$} \molenohammer)}               & -69.34               & -56.02      & -5.61 \footnotesize{(\textcolor{red}{$\times 1.08$} \molenohammer)}                \\
JTT                       & 76.33                & -26.40  \footnotesize{(\textcolor{red}{$\times 1.04$} \molenohammer)}               & +32   \footnotesize{(\textcolor{red}{$\times 1.06$} \molenohammer)}               & -69.48             & -56.30      & -5.55 \footnotesize{(\textcolor{red}{$\times 1.07$} \molenohammer)}             \\
EIIL                        & 71.55                & -33.48 \footnotesize{(\textcolor{red}{$\times 1.31$} \molenohammer)}               & +24                  & -66.04               & -61.35 \footnotesize{(\textcolor{red}{$\times 1.09$} \molenohammer)}     & -6.42 \footnotesize{(\textcolor{red}{$\times 1.24$} \molenohammer)}              \\
DebiAN                     & 76.33                & -26.40  \footnotesize{(\textcolor{red}{$\times 1.04$} \molenohammer)}          & +36  \footnotesize{(\textcolor{red}{$\times 1.20$} \molenohammer)}                & -69.37               & -56.29      & -5.53 \footnotesize{(\textcolor{red}{$\times 1.07$} \molenohammer)}               \\ \midrule \midrule
\textbf{LLE (ours)}        & 76.25                & -6.18                & \textbf{+10}         & \textbf{-61.00}      & -54.89               & \textbf{-3.82}       \\ \bottomrule
\end{tabular}
\end{adjustbox}
\caption{\textbf{Existing methods fail to combat multiple shortcuts by amplifying at least one shortcut relative to ERM on ImageNet.} All models use ResNet-50 with last layer re-training~\cite{kirichenko2022Last}.}
\label{tab:imagenet_results}
\vspace{-4mm}
\end{table}

\subsection{Results: Mitigation Methods}
\label{subsec:mitigation_methods_results}

\noindent \textbf{Results: Standard Augmentation and Regularization (Category 1)} \quad We first show the results of methods using augmentation and regularization without using inductive biases of shortcuts.
On UrbanCars (\cref{tab:urbancars_unsupervised_results}), we observed that CutMix and Cutout amplify the background shortcut with a larger BG Gap relative to ERM. AugMix increases the reliance on the CoObj shortcut with a larger CoObj Gap (\ie, -12.2\%) compared to ERM. Although Mixup and SD do not produce Whac-A-Mole results, they only yield marginal improvement or can only mitigate one shortcut well.
On ImageNet, the results in \cref{tab:imagenet_results} show that all approaches amplify at least one shortcut. For instance, AugMix achieves a worse Carton Gap to amplify the watermark shortcut compared to ERM. For CutMix, we again observe that it amplifies the BG shortcut on ImageNet. We show more results of CutMix and analyze its background shortcut reliance in \cref{appx:sec:cutmix_amp_bg_shortcut}.
\begin{mybox}
    \textbf{Takeaway}: Standard augmentation and regularization methods can mitigate some shortcuts (\eg, texture) \molehammer{} but amplify others \molenohammer{}.
\end{mybox}

\noindent \textbf{Results: Targeted Augmentation for Mitigating Shortcuts (Category 2)} \quad Further, we benchmark methods using data augmentation to mitigate a specific shortcut. Compared to methods in category 1, augmentations here use stronger inductive biases about the shortcut by modifying the shortcut visual cue. On UrbanCars, although CF+F Aug achieves good results for the CoObj shortcut, it amplifies the BG shortcut. On ImageNet, texture and background augmentation improve the reliance on the watermark shortcut, which can be explained by the retained or even increased saliency of the watermark in \cref{fig:teaser_in_w} and Appendix's \cref{appx:fig:style_transfer_watermark,appx:fig:bg_aug_watermark}.
\begin{mybox}
    \textbf{Takeaway}: Augmentations tackling a specific type of shortcut~\molehammer{} (\eg, style transfer for texture shortcut) can amplify other shortcuts~\molenohammer{} (\eg, watermark).
\end{mybox}

\begin{table}[t]
\centering
\begin{adjustbox}{width=\linewidth}
\begin{tabular}{@{}lllclll@{}}
\toprule
    & \multicolumn{2}{c}{shortcut label} & \multicolumn{1}{c|}{}         & \multicolumn{3}{c}{shortcut reliance} \\ \cmidrule(lr){2-3} \cmidrule(l){5-7}
    & Train                     & Val    & \multicolumn{1}{c|}{I.D. Acc} & BG Gap $\uparrow$  & CoObj Gap $\uparrow$ & BG+CoObj Gap $\uparrow$  \\ \midrule
\textcolor{gray}{ERM}    & \textcolor{gray}{\xmark}      & \textcolor{gray}{BG+CoObj}          & \textcolor{gray}{97.6}       & \textcolor{gray}{-15.3}          & \textcolor{gray}{-11.2}          & \textcolor{gray}{-69.2}          \\
gDRO              & BG+CoObj          & BG+CoObj                         & 91.6                                                    & -10.9          & -3.6           & -16.4          \\
DI                & BG+CoObj          & BG+CoObj                         & 89.0                                                    & \textbf{-2.2}  & -1.0  & \textbf{+0.4}  \\
SUBG              & BG+CoObj          & BG+CoObj                         & 71.1                                                    & -4.7           & \textbf{-0.3}           & -6.3           \\
DFR               & BG+CoObj          & BG+CoObj                         & 89.7                                                    & -10.7           & -6.9          & -45.2          \\ \midrule
\textcolor{gray}{ERM}   & \textcolor{gray}{\xmark}    & \textcolor{gray}{BG}      & \textcolor{gray}{97.8}       & \textcolor{gray}{-14.6}          & \textcolor{gray}{-11.3}          & \textcolor{gray}{-68.5}          \\
gDRO              & BG \molehammer                & BG                               & 96.0                                                    & -4.2            & -26.9 (\textcolor{red}{$\times$2.39} \molenohammer)         & -56.5          \\
DI                & BG \molehammer               & BG                               & 94.7                                                    & +2.2            & -27.0  (\textcolor{red}{$\times$2.40} \molenohammer)        & -25.2          \\
SUBG              & BG \molehammer               & BG                               & 92.6                                                    & +1.3            & -36.4 (\textcolor{red}{$\times$3.24} \molenohammer)          & -35.8          \\
DFR               & BG \molehammer               & BG                               & 97.4                                                    & -9.8          & -13.6 (\textcolor{red}{$\times$1.21} \molenohammer)         & -58.9          \\ \midrule
\textcolor{gray}{ERM}     & \textcolor{gray}{\xmark} & \textcolor{gray}{CoObj}    & \textcolor{gray}{97.6}    & \textcolor{gray}{-15.4}          & \textcolor{gray}{-11.0}          & \textcolor{gray}{-68.8}          \\
gDRO              & CoObj \molehammer            & CoObj                            & 95.7                                                    & -31.4 (\textcolor{red}{$\times$2.03} \molenohammer)         & -0.5            & -54.9          \\
DI                & CoObj \molehammer            & CoObj                            & 94.2                                                    & -36.1 (\textcolor{red}{$\times$2.34} \molenohammer)         & +2.8            & -35.8          \\
SUBG              & CoObj \molehammer            & CoObj                            & 93.1                                                    & -60.2  (\textcolor{red}{$\times$3.90} \molenohammer)        & +2.5            & -62.4          \\
DFR               & CoObj \molehammer            & CoObj                            & 97.4                                                    & -19.1  (\textcolor{red}{$\times$1.24} \molenohammer)        & -8.6           & -64.9          \\
 \bottomrule
\end{tabular}
\end{adjustbox}
\caption{\textbf{Methods using shortcut labels (category 3) amplify the unlabeled shortcut when mitigating the labeled shortcut on UrbanCars.} \molehammer: mitigate a shortcut, \eg, using shortcut labels.}
\label{tab:urbancars_results}
\vspace{-4mm}
\end{table}

\noindent \textbf{Results: Using Shortcut Labels (Category 3)} \quad
Then, we show the results of methods using shortcut labels on UrbanCars in \cref{tab:urbancars_results}. Methods can mitigate multiple shortcuts when labels of both shortcuts are used (\cf first section in \cref{tab:urbancars_results}). However, when using labels of either shortcut, which is the typical situation for in-the-wild datasets where shortcut labels are incomplete, they exhibit a higher performance gap in the other shortcut relative to ERM. \Eg, when only using the CoObj labels, models achieve poorer BG Gap results.
\begin{mybox}
    \textbf{Takeaway}: Methods using shortcut labels mitigate the labeled shortcut \molehammer{} but amplifies the unlabeled one \molenohammer.
\end{mybox}

\noindent \textbf{Results: Inferring Pseudo Shortcut Labels (Category 4)} \quad The Whac-A-Mole problem of methods using shortcut labels motivates us to study whether the problem can be solved by inferring pseudo labels of multiple shortcuts.
Here we analyze the results of LfF, JTT, EIIL, and DebiAN.
Their key idea is based on ERM's training dynamics of learning different visual cues.
LfF infers soft shortcut labels by assuming that the shortcut is learned earlier.
Similarly, JTT and EIIL use an under-trained ERM trained with E epochs as the reference model to infer pseudo shortcut labels. We use E=1 and E=2 for JTT and EIIL.
Instead of using a fixed reference model, DebiAN jointly trains the reference and mitigation models.
The results in \cref{tab:urbancars_unsupervised_results} show that LfF, JTT (E=1), and EIIL (E=1) still exhibit Whac-A-Mole results by achieving a larger CoObj Gap than ERM. On the other hand, JTT (E=2) and EIIL (E=2) also show the Whac-A-Mole results by achieving larger BG Gap than ERM.
On ImageNet, we observe Whac-A-Mole results produced by LfF, JTT, EIIL, and DebiAN in \cref{tab:imagenet_results}.

\begin{figure}[t]
  \centering
   \includegraphics[width=\linewidth]{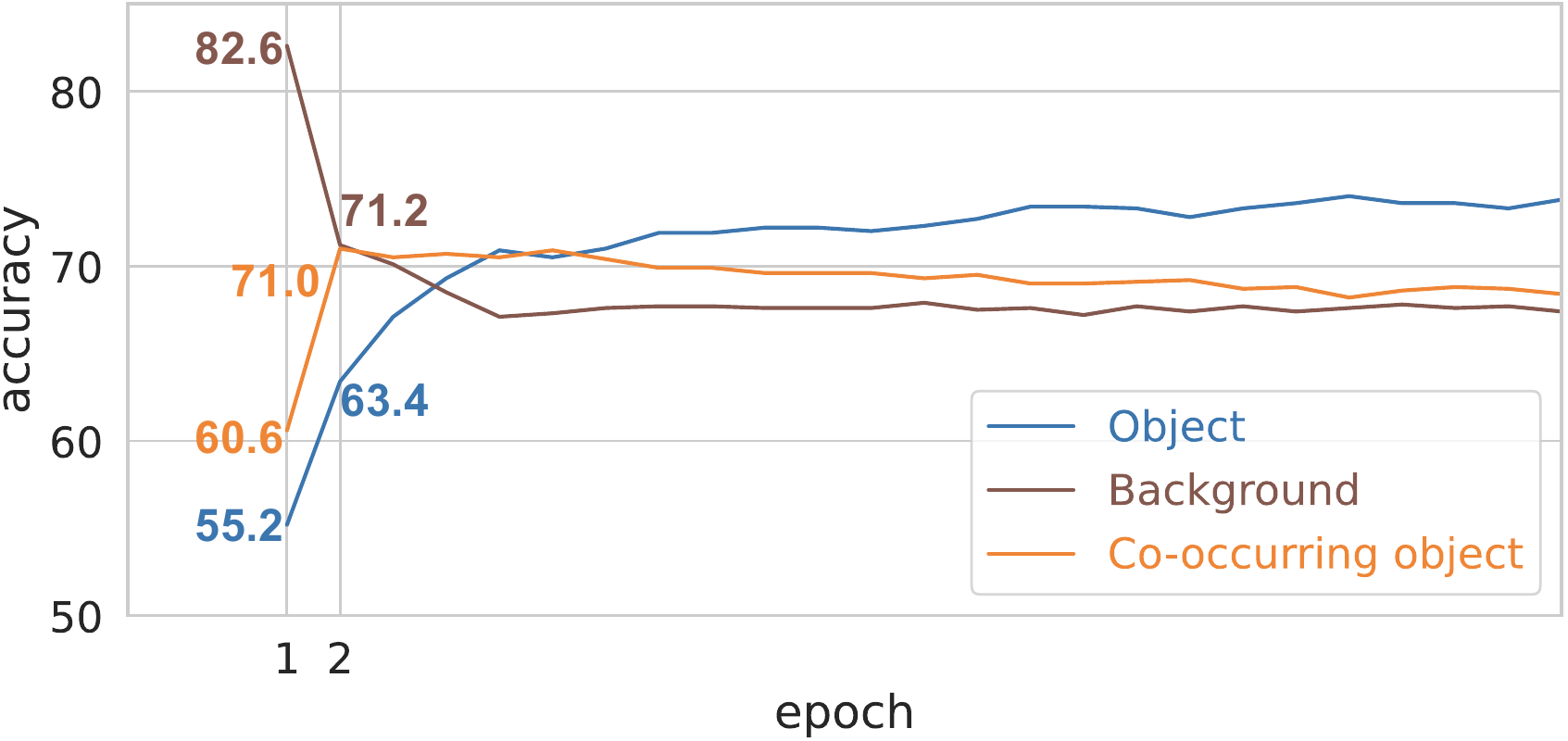}
   \caption{On UrbanCars, \textbf{ERM learns BG and CoObj shortcuts at different training epochs, making it difficult to infer pseudo labels (category 4) of multiple shortcuts from ERM.}
   }
   \label{fig:cues_acc_along_epochs}
  \vspace{-4mm}
\end{figure}

\textbf{To investigate the reason for their Whac-A-Mole results, we analyze the \textit{training dynamics of ERM}.} In \cref{fig:cues_acc_along_epochs}, we plot the accuracy of three visual cues---object (\ie, car body type), background, and co-occurring object on the validation set. The accuracy is computed based on ERM's $\{\texttt{urban}, \texttt{country}\}$ predictions against labels of object, BG, and CoObj. We observe a Whac-A-Mole game in ERM's training. At epoch 1, ERM mainly predicts the background (82.6\%), suggesting that the background shortcut is learned first. Thus, LfF, JTT (E=1), and EIIL (E=1) can infer the BG shortcut labels well to amplify the CoObj shortcut.
As the training continues to epoch 2, the reliance on the BG shortcut decreases (82.6\% to 71.2\%), but the reliance on the CoObj shortcut is increased (60.6\% to 71.8\%). It renders JTT (E=2) and EIIL (E=2) better infer CoObj shortcut labels, which, in turn, amplifies the BG shortcut.
\begin{mybox}
    \textbf{Takeaway}: Methods inferring pseudo shortcut labels still amplify shortcuts \molenohammer{} because ERM learns different shortcuts \textit{asynchronously} during training, making it hard to infer labels of all shortcuts \molehammer{} for mitigation.
\end{mybox}

\subsection{Results: Self-Supervised \& Foundation Models}
\label{subsec:self_sup_foundation_results}

On ImageNet, we further benchmark self-supervised pretraining methods, \ie, MoCov3~\cite{chen2021IEEECVFInt.Conf.Comput.Vis.ICCVEmpirical}, MAE~\cite{he2022IEEECVFConf.Comput.Vis.PatternRecognit.CVPRMasked}, SEER~\cite{goyal2022Vision}. We also benchmark foundation models that use extra training data, \ie, Uniform Soup~\cite{wortsman2022Int.Conf.Mach.Learn.Model}, Greedy Soup~\cite{wortsman2022Int.Conf.Mach.Learn.Model}, CLIP~\cite{radford2021Int.Conf.Mach.Learn.Learning}, SEER~\cite{goyal2022Vision}, and SWAG~\cite{singh2022IEEECVFConf.Comput.Vis.PatternRecognit.CVPRRevisiting}.
The results in \cref{tab:imagenet_results_larger_arch} show that many of them fail to mitigate multiple shortcuts jointly.
Regarding self-supervised methods, MoCov3 achieves worse results on all three shortcuts, and MAE achieves a worse SIN Gap for the texture shortcut relative to ERM.
Regarding foundation models, although SWAG with linear probing (LP) achieves a much better IN-R Gap (-19.79\%), it also has a stronger reliance on the background in BG Gap compared to ERM.
Similarly, SEER, Uniform Soup, and Greedy Soup mitigate the watermark shortcut but amplify the background shortcut.
When using ViT-L, although CLIP with zero-shot transfer does not produce Whac-A-Mole results, they do not fully close the performance gap. Besides, they also show much lower IN-1k accuracy than other foundation models. We show results using other architectures in \cref{appx:subsec:imagenet_lle_other_arch}.
\begin{mybox}
    \textbf{Takeaway}: Self-supervised and foundation models can mitigate some shortcuts \molehammer{} but amplify others \molenohammer.
\end{mybox}

\begin{table}[t]
\centering
\begin{adjustbox}{width=\linewidth}
\begin{tabular}{@{}lllllll@{}}
\toprule
 &                            & \multicolumn{5}{|c}{shortcut reliance}                                                                                                               \\ \cmidrule(l){3-7}
 & \multicolumn{1}{c|}{}      & \multicolumn{2}{c|}{Watermark}                             & \multicolumn{2}{c|}{Texture}                          & \multicolumn{1}{c}{Background} \\
 & \multicolumn{1}{l|}{IN-1k} & IN-W $\uparrow$ & \multicolumn{1}{l|}{Carton $\downarrow$} & SIN $\uparrow$ & \multicolumn{1}{l|}{IN-R $\uparrow$} & IN-9 $\uparrow$                \\
 & \multicolumn{1}{l|}{}      & Gap             & \multicolumn{1}{l|}{Gap}                 & Gap            & \multicolumn{1}{l|}{Gap}             & Gap                            \\  \midrule
\textit{arch: RG-32gf}                   &                    &          &           &          &         &                 \\
\textcolor{gray}{ERM}                  & \textcolor{gray}{80.88}                   & \textcolor{gray}{-14.15}           & \textcolor{gray}{+32}          & \textcolor{gray}{\textbf{-69.27}}          & \textcolor{gray}{-52.43}          & \textcolor{gray}{\textbf{-6.40}}                  \\
SEER \scriptsize{(FT,IG-1B)}                 & 83.35          & \textbf{-6.50}   & \textbf{+18} & -73.04 \scriptsize{(\textcolor{red}{$\times 1.05$} \molenohammer)}         & \textbf{-50.42}          & -7.14 \scriptsize{(\textcolor{red}{$\times 1.11$} \molenohammer)}                          \\ \midrule
\textit{arch: ViT-B/32}                   &                    &          &           &          &         &                 \\
\textcolor{gray}{ERM}                & \textcolor{gray}{75.92}                  & \textcolor{gray}{-8.71}            & \textcolor{gray}{+34}          & \textcolor{gray}{\textbf{-57.16}}          & \textcolor{gray}{-49.45}          & \textcolor{gray}{\textbf{-6.86}}                 \\
Uniform Soup \scriptsize{(FT,WIT)}      & 79.96         & -7.90   & +24 & -59.67 \scriptsize{(\textcolor{red}{$\times 1.04$} \molenohammer)} & \textbf{-27.51} & -7.78  \scriptsize{(\textcolor{red}{$\times 1.13$} \molenohammer)}      \\
Greedy Soup \scriptsize{(FT,WIT)}      & 81.01         & \textbf{-6.47}   & \textbf{+16} & -59.61 \scriptsize{(\textcolor{red}{$\times 1.04$} \molenohammer)} & -30.01 & -7.21  \scriptsize{(\textcolor{red}{$\times 1.05$} \molenohammer)}      \\\midrule
\textit{arch: ViT-B/16}                  &                    &          &           &          &         &                 \\
\textcolor{gray}{ERM}        & \textcolor{gray}{81.07}          & \textcolor{gray}{-6.69}   & \textcolor{gray}{+26}          & \textcolor{gray}{-62.60}           & \textcolor{gray}{-50.36}          & \textcolor{gray}{-5.36}                 \\
SWAG \scriptsize{(LP,IG-3.6B)}       & 81.89                   & -7.76  \scriptsize{(\textcolor{red}{$\times 1.16$} \molenohammer)}           & +18 & -67.33 \scriptsize{(\textcolor{red}{$\times 1.08$} \molenohammer)} & \textbf{-19.79} & -10.39  \scriptsize{(\textcolor{red}{$\times 1.94$} \molenohammer)}                         \\
SWAG \scriptsize{(FT,IG-3.6B)}       & 85.29                   & -5.43             & +24 & -66.99 \scriptsize{(\textcolor{red}{$\times 1.07$} \molenohammer)} & -29.55 & -4.44                           \\
MoCov3 \scriptsize{(LP)}           & 76.65                   & -16.0  \scriptsize{(\textcolor{red}{$\times 2.39$} \molenohammer)}            & +22 & -63.36 \scriptsize{(\textcolor{red}{$\times 1.01$} \molenohammer)} & -56.86 \scriptsize{(\textcolor{red}{$\times 1.12$} \molenohammer)} & -7.80 \scriptsize{(\textcolor{red}{$\times 1.45$} \molenohammer)}                           \\
MAE \scriptsize{(FT)}       & 83.72                   & -4.60             & +24 & -65.20 \scriptsize{(\textcolor{red}{$\times 1.04$} \molenohammer)} & -47.10 & -4.45                           \\
MAE+\textbf{LLE (ours)}         & 83.68                   & \textbf{-2.48}             & \textbf{+6} & \textbf{-58.78} & -44.96 & \textbf{-3.70}                           \\ \midrule
\textit{arch: ViT-L/16 or 14}                   &                    &          &           &          &         &                 \\
\textcolor{gray}{ERM}          & \textcolor{gray}{79.65}          & \textcolor{gray}{-6.14}   & \textcolor{gray}{+34}          & \textcolor{gray}{-61.43}           & \textcolor{gray}{-53.17}          & \textcolor{gray}{-6.50}                 \\
SWAG \scriptsize{(LP,IG-3.6B)}          & 85.13                  & -5.73           & \textbf{+6}  & -60.26          & -10.17           & -7.26 \scriptsize{(\textcolor{red}{$\times 1.12$} \molenohammer)}                         \\
SWAG \scriptsize{(FT,IG-3.6B)}         & 88.07         & -3.16           & +20          & -63.45 \scriptsize{(\textcolor{red}{$\times 1.03$} \molenohammer)}         & -12.29           & -2.92                 \\
CLIP \scriptsize{(zero-shot,WIT)}      & 76.57            & -4.47           & +12          & -61.27 & \textbf{-6.26}  & -3.68                          \\
CLIP \scriptsize{(zero-shot,LAION)}      & 72.77            & -4.94           & +12          & -56.85 & -8.43  & -4.54                          \\
MAE \scriptsize{(FT)}         & 85.95         & -4.36           & +22          & -62.48 \scriptsize{(\textcolor{red}{$\times 1.02$} \molenohammer)}         & -36.46          & -3.53                          \\
MAE+\textbf{LLE (ours)}       & 85.84                  & \textbf{-1.74}  & +12 & \textbf{-56.32} & -34.64 & \textbf{-2.77}                 \\
\bottomrule
\end{tabular}
\end{adjustbox}
\caption{On ImageNet, many \textbf{self-supervised and foundation models amplify shortcuts}, whereas LLE mitigates multiple shortcuts jointly. ($\cdot$): transfer learning (and extra data).
}
\label{tab:imagenet_results_larger_arch}
\vspace{-6mm}
\end{table}

\subsection{Results: Last Layer Ensemble (LLE)}
\label{subsec:lle_results}

We show that our Last Layer Ensemble (LLE) can better tackle multi-shortcut mitigation. LLE mitigates shortcuts via a set of data augmentations. Specifically,  we augment background (BG) and co-occurring object (CoObj) by swapping BG and CoObj across target classes on UrbanCars (details in \cref{appx:subsec:lle_details}). On ImageNet, we use watermark augmentation (WMK Aug), style transfer~\cite{geirhos2019Int.Conf.Learn.Represent.ImageNettrained} (TXT Aug), and background augmentation~\cite{xiao2021Int.Conf.Learn.Represent.Noise,ryali2021Characterizing} (BG Aug) for watermark, texture, and background shortcuts, respectively.

The results on UrbanCars in \cref{tab:urbancars_unsupervised_results} show that LLE beats all other methods in BG Gap and BG+CoObj Gap metrics and achieves second best CoObj Gap to CF+F Aug, a method amplifies the background shortcut.
The results of ImageNet with ResNet-50 are in \cref{tab:imagenet_results}. LLE achieves the best multi-shortcut mitigation results in Carton Gap, SIN Gap, and IN-9 Gap.
Regarding IN-W Gap and IN-R Gap, LLE achieves better results than ERM. \Ie, no Whac-A-Mole problems.
On ImageNet, we further use MAE as the feature extractor, and the results on ImageNet are in \cref{tab:imagenet_results_larger_arch}.
LLE achieves the best results in IN-W Gap, SIN Gap, and IN-9 Gap. LLE also achieves the best results in the remaining metrics comparing to methods not using extra pretraining data.

\noindent \textbf{Ablation Study} \quad In \cref{tab:lle_ablation_imagenet}, we show the ablation study of LLE: (1) w/o ensemble: training a single last layer. (2) AugMix (without ensemble): based on (1) and use JS divergence in AugMix to improve the invariance across augmentations. (3) w/o dist cls.: remove \textit{domain shift classifier} and directly take the mean over the output of ensemble classifiers. Except for IN-R Gap, the full model achieves better results in all other metrics. Although the w/o ensemble achieves a better IN-R Gap, it suffers from reliance on other shortcuts.

\section{Related Work}

\noindent \textbf{Group Shift Datasets} \quad
Most previous works use single-shortcut datasets~\cite{sagawa2020Int.Conf.Learn.Represent.Distributionally,arjovsky2020Invariant,nam2020Adv.NeuralInf.Process.Syst.Learning,kim2021IEEECVFInt.Conf.Comput.Vis.ICCVBiaSwap,liu2015IEEEInt.Conf.Comput.Vis.ICCVDeep,liang2022Int.Conf.Learn.Represent.MetaShift,he2021PatternRecognitionNonI,koh2021Proc.38thInt.Conf.Mach.Learn.WILDS} to benchmark group shift robustness~\cite{sagawa2020Int.Conf.Learn.Represent.Distributionally}.
Although \cite{seo2022IEEECVFConf.Comput.Vis.PatternRecognit.CVPRUnsupervised,bao2021Int.Conf.Mach.Learn.Predict,zhao2022Submitt.Int.Conf.Learn.Represent.Scaling} use labels of multiple attributes~\cite{liu2015IEEEInt.Conf.Comput.Vis.ICCVDeep} for evaluation, there lacks a sanity check on whether the selected attributes are learned as spurious shortcuts.
\cite{shrestha2022IEEECVFWinterConf.Appl.Comput.Vis.WACVInvestigation,li2022Eur.Conf.Comput.Vis.ECCVDiscovera} create MNIST-based~\cite{lecun1998Proc.IEEEGradientbased} synthetic datasets with multiple shortcuts, where the shortcuts are unrealistic.
In contrast, our UrbanCars dataset is more photo-realistic and contains commonly seen shortcuts.
Besides, our ImageNet-W dataset better evaluates shortcut mitigation on the large-scale and real-world ImageNet dataset.

\noindent \textbf{OOD Datasets of ImageNet} \quad While many models achieve great performance on ImageNet~\cite{deng2009IEEEConf.Comput.Vis.PatternRecognit.CVPRImageNet}, they suffer under various distributional shifts, \eg, corruption~\cite{hendrycks2019Int.Conf.Learn.Represent.Benchmarking}, sketches~\cite{wang2019Adv.NeuralInf.Process.Syst.Learning}, rendition~\cite{hendrycks2021IEEECVFInt.Conf.Comput.Vis.ICCVMany}, texture~\cite{geirhos2019Int.Conf.Learn.Represent.ImageNettrained}, background~\cite{xiao2021Int.Conf.Learn.Represent.Noise}, or unknown distributional shifts~\cite{hendrycks2021IEEECVFConf.Comput.Vis.PatternRecognit.CVPRNatural,recht2019Proc.36thInt.Conf.Mach.Learn.ImageNet}.
In this work, we construct ImageNet-W, where SoTA vision models rely on our newly discovered watermark shortcut.

\noindent \textbf{Shortcut Mitigation and Improving OOD Robustness} \quad
To address the shortcut learning problem~\cite{geirhos2020NatMachIntellShortcut}, \cite{sagawa2020Int.Conf.Learn.Represent.Distributionally,wang2020IEEECVFConf.Comput.Vis.PatternRecognit.CVPRFairness,idrissi2022Conf.CausalLearn.Reason.Simple} use shortcut labels for mitigation.
With only knowledge of the shortcut type, \cite{wang2019Int.Conf.Learn.Represent.Learning,bahng2020Int.Conf.Mach.Learn.Learning} use architectural inductive biases. \cite{geirhos2019Int.Conf.Learn.Represent.ImageNettrained,xiao2021Int.Conf.Learn.Represent.Noise,ryali2021Characterizing} use augmentation and \cite{kirichenko2022Last,izmailov2022Adv.NeuralInf.Process.Syst.Feature} re-trains the last layer for mitigation.
Without knowledge of shortcut types, \cite{nam2020Adv.NeuralInf.Process.Syst.Learning,sohoni2020Adv.NeuralInf.Process.Syst.No,seo2022IEEECVFConf.Comput.Vis.PatternRecognit.CVPRUnsupervised,liu2021Int.Conf.Mach.Learn.Just,creager2021Int.Conf.Mach.Learn.Environment,ahmed2021Int.Conf.Learn.Represent.Systematic,zhang2022Int.Conf.Mach.Learn.Rich,li2022Eur.Conf.Comput.Vis.ECCVDiscovera} infer pseudo shortcut labels, which is theoretically impossible~\cite{lin2022Adv.NeuralInf.Process.Syst.ZINa}, and we show that they struggle to mitigate multiple shortcuts.
Other works suggest that self-supervised pretraining~\cite{he2022IEEECVFConf.Comput.Vis.PatternRecognit.CVPRMasked,kim2022Adv.NeuralInf.Process.Syst.Learning} and foundation models~\cite{bommasani2022Opportunities,goyal2022Vision,radford2021Int.Conf.Mach.Learn.Learning,ilharco2021OpenCLIP,goyal2022Vision,wortsman2022IEEECVFConf.Comput.Vis.PatternRecognit.CVPRRobust,wortsman2022Int.Conf.Mach.Learn.Model} improve OOD robustness. We show that many of them suffer from the Whac-A-Mole problem or struggle to close performance gaps.

\begin{table}[t]
  \centering
  \begin{adjustbox}{width=\linewidth}
  \begin{tabular}{@{}lcccccc@{}}
  \toprule
                        & \multicolumn{1}{c|}{}      & \multicolumn{5}{c}{Shortcut Reliance}                                                                               \\ \cmidrule(l){3-7}
                     & \multicolumn{1}{c|}{}      & \multicolumn{2}{c|}{Watermark}                   & \multicolumn{2}{c|}{Texture}                    & Background     \\
                            & \multicolumn{1}{c|}{IN-1k} & IN-W Gap $\uparrow$      & \multicolumn{1}{c|}{Carton Gap $\downarrow$} & SIN Gap $\uparrow$         & \multicolumn{1}{c|}{IN-R Gap $\uparrow$} & IN-9 Gap $\uparrow$      \\ \midrule
  w/o ensemble          & 76.03                      & -6.71          & +18                             & -66.81          & \textbf{-52.55}               & -5.08          \\
  AugMix            & 75.17                      & -7.27          & +22                             & -66.33          & -56.38               & -5.38          \\
  w/o dist. cls.        & 75.82                      & -17.77         & +36                             & -66.45          & -53.58                        & -4.81          \\
  \textbf{LLE (full model)}      & \textbf{76.25}             & \textbf{-6.18} & \textbf{+10}                    & \textbf{-61.20} & -54.89                        & \textbf{-3.82} \\ \bottomrule
  \end{tabular}
  \end{adjustbox}
  \caption{Ablation study of Last Layer Ensemble on ImageNet.}
  \label{tab:lle_ablation_imagenet}
  \vspace{-6mm}
\end{table}

\section{Conclusion}

We propose novel benchmarks to evaluate multi-shortcut mitigation.
The results show that state-of-the-art models, ranging from shortcut mitigation methods to foundation models, fail to mitigate multiple shortcuts in a Whac-A-Mole game.
To tackle this open challenge, we propose Last Layer Ensemble method to mitigate multiple shortcuts jointly.
We leave to future work for shortcut mitigation without knowledge of shortcut types. Another promising future direction is to provide a theoretical analysis of the Whac-A-Mole phenomenon.
Finally, we call for discarding the tenuous single-shortcut assumption and hope our work can inspire future research into the overlooked challenge of multi-shortcut mitigation.

\vspace{5pt}
\noindent \textbf{Acknowledgment} \quad This work has been partially supported by the National Science Foundation (NSF) under Grant 1909912 and by the Center of Excellence in Data Science, an Empire State Development-designated Center of Excellence. The article solely reflects the opinions and conclusions of its authors but not the funding agents.

\clearpage
\nocite{FlaticonWhack} %

\balance
{\small
\bibliographystyle{ieee_fullname}
\bibliography{ref}

\begin{thebibliography}{10}
\bibitem[Fla()]{FlaticonWhack}
Whack {{A Mole}} image is obtained from {{Flaticon}}.com.
\end{thebibliography}
}

\clearpage

\onecolumn

\appendix

\section*{Appendix}

\section{More Details of Datasets}

\subsection{UrbanCars Details}
\label{appx:subsec:urbancars_details}

Here we present more details of the UrbanCars dataset.

\paragraph{Number of Images}
Regarding the number of images, each target class contains 4000 images in the training set, \ie, 8000 images in total. That is, our training set is balanced regarding the target label and only imbalanced with shortcut labels.
Therefore, UrbanCars does not have a target class imbalance issue~\cite{idrissi2022Conf.CausalLearn.Reason.Simple} in Waterbirds dataset~\cite{sagawa2020Int.Conf.Learn.Represent.Distributionally}, where 76.8\% of images are waterbird, and 23.3\% of images are landbird.
In validation and testing sets of UrbanCars, each split contains 500 images.

\paragraph{Data Annotation} As mentioned in \cref{subsec:urbancars}, each image is annotated with three image-level labels---car body type, background, and co-occurring object. Besides, following Waterbirds~\cite{sagawa2020Int.Conf.Learn.Represent.Distributionally} dataset, the dataset also contains the mask annotation of the car object and the co-occurring object, which enables shortcut mitigation via targeted augmentation (category 2), \ie, CF+F Aug~\cite{chang2021IEEECVFConf.Comput.Vis.PatternRecognit.CVPRRobust} (\cf \cref{appx:subsec:benchmark_methods_details}) and our proposed LLE approach (\cf \cref{appx:subsec:lle_details}).

\paragraph{Details of Data Construction} Here, we present the details of collecting the data from source datasets based on three visual cues---main object (\ie, car), background shortcut, and co-occurring object shortcut.

First, to obtain car images, we use MaskFormer~\cite{cheng2021Adv.NeuralInf.Process.Syst.PerPixel} pretrained on MS-COCO~\cite{lin2014Eur.Conf.Comput.Vis.ECCVMicrosoft} dataset's panoptic segmentation~\cite{kirillov2019IEEECVFConf.Comput.Vis.PatternRecognit.CVPRPanoptic} task to segment cars from Stanford Cars~\cite{krause2013IEEEInt.Conf.Comput.Vis.Workshop3D} dataset. In each image from Stanford Cars, we choose the predicted car instance mask that has the largest IoU with the bounding box annotation provided in Stanford Cars. After segmentation, we run MaskFormer on foreground-only images to detect humans. Images with humans detected are filtered out.

When pasting the car object to the background, we first compute its square bounding box, which is the bounding box whose side length is the longer side of the actual bounding box of the car object based on the predicted segmentation mask. Then, we resize the square bounding box such that the side length is 50\% of the final image size, which is smaller than the size of the car object.

We merge the original 196 classes in Stanford Cars into urban cars (\eg, sedan, hatchback, \etc) and country cars (\eg, pickup truck, van, \etc).  The mapping from the original 196 classes in Stanford Cars to \textit{urban cars} and \textit{country cars} is as follows:
\begin{itemize}[noitemsep,topsep=0pt]
    \item \textit{urban} cars: Acura RL Sedan 2012, Acura TL Sedan 2012, Acura TL Type-S 2008, Acura TSX Sedan 2012, Acura Integra Type R 2001, Acura ZDX Hatchback 2012, Aston Martin V8 Vantage Coupe 2012, Aston Martin Virage Convertible 2012, Aston Martin Virage Coupe 2012, Audi RS 4 Convertible 2008, Audi A5 Coupe 2012, Audi TTS Coupe 2012, Audi R8 Coupe 2012, Audi V8 Sedan 1994, Audi 100 Sedan 1994, Audi 100 Wagon 1994, Audi TT Hatchback 2011, Audi S6 Sedan 2011, Audi S5 Convertible 2012, Audi S5 Coupe 2012, Audi S4 Sedan 2012, Audi S4 Sedan 2007, Audi TT RS Coupe 2012, BMW ActiveHybrid 5 Sedan 2012, BMW 1 Series Convertible 2012, BMW 1 Series Coupe 2012, BMW 3 Series Sedan 2012, BMW 3 Series Wagon 2012, BMW 6 Series Convertible 2007, BMW M3 Coupe 2012, BMW M5 Sedan 2010, BMW M6 Convertible 2010, BMW Z4 Convertible 2012, Bentley Continental Supersports Conv. Convertible 2012, Bentley Arnage Sedan 2009, Bentley Mulsanne Sedan 2011, Bentley Continental GT Coupe 2012, Bentley Continental GT Coupe 2007, Bentley Continental Flying Spur Sedan 2007, Bugatti Veyron 16.4 Convertible 2009, Bugatti Veyron 16.4 Coupe 2009, Buick Regal GS 2012, Buick Verano Sedan 2012, Cadillac CTS-V Sedan 2012, Chevrolet Corvette Convertible 2012, Chevrolet Corvette ZR1 2012, Chevrolet Corvette Ron Fellows Edition Z06 2007, Chevrolet Camaro Convertible 2012, Chevrolet Impala Sedan 2007, Chevrolet Sonic Sedan 2012, Chevrolet Cobalt SS 2010, Chevrolet Malibu Hybrid Sedan 2010, Chevrolet Monte Carlo Coupe 2007, Chevrolet Malibu Sedan 2007, Chrysler Sebring Convertible 2010, Chrysler 300 SRT-8 2010, Chrysler Crossfire Convertible 2008, Chrysler PT Cruiser Convertible 2008, Daewoo Nubira Wagon 2002, Dodge Caliber Wagon 2012, Dodge Caliber Wagon 2007, Dodge Magnum Wagon 2008, Dodge Challenger SRT8 2011, Dodge Charger Sedan 2012, Dodge Charger SRT-8 2009, Eagle Talon Hatchback 1998, FIAT 500 Abarth 2012, FIAT 500 Convertible 2012, Ferrari FF Coupe 2012, Ferrari California Convertible 2012, Ferrari 458 Italia Convertible 2012, Ferrari 458 Italia Coupe 2012, Fisker Karma Sedan 2012, Ford Mustang Convertible 2007, Ford GT Coupe 2006, Ford Focus Sedan 2007, Ford Fiesta Sedan 2012, Geo Metro Convertible 1993, Honda Accord Coupe 2012, Honda Accord Sedan 2012, Hyundai Veloster Hatchback 2012, Hyundai Sonata Hybrid Sedan 2012, Hyundai Elantra Sedan 2007, Hyundai Accent Sedan 2012, Hyundai Genesis Sedan 2012, Hyundai Sonata Sedan 2012, Hyundai Elantra Touring Hatchback 2012, Hyundai Azera Sedan 2012, Infiniti G Coupe IPL 2012, Jaguar XK XKR 2012, Lamborghini Reventon Coupe 2008, Lamborghini Aventador Coupe 2012, Lamborghini Gallardo LP 570-4 Superleggera 2012, Lamborghini Diablo Coupe 2001, Lincoln Town Car Sedan 2011, MINI Cooper Roadster Convertible 2012, Maybach Landaulet Convertible 2012, McLaren MP4-12C Coupe 2012, Mercedes-Benz 300-Class Convertible 1993, Mercedes-Benz C-Class Sedan 2012, Mercedes-Benz SL-Class Coupe 2009, Mercedes-Benz E-Class Sedan 2012, Mercedes-Benz S-Class Sedan 2012, Mitsubishi Lancer Sedan 2012, Nissan Leaf Hatchback 2012, Nissan Juke Hatchback 2012, Nissan 240SX Coupe 1998, Plymouth Neon Coupe 1999, Porsche Panamera Sedan 2012, Rolls-Royce Phantom Drophead Coupe Convertible 2012, Rolls-Royce Ghost Sedan 2012, Rolls-Royce Phantom Sedan 2012, Scion xD Hatchback 2012, Spyker C8 Convertible 2009, Spyker C8 Coupe 2009, Suzuki Aerio Sedan 2007, Suzuki Kizashi Sedan 2012, Suzuki SX4 Hatchback 2012, Suzuki SX4 Sedan 2012, Tesla Model S Sedan 2012, Toyota Camry Sedan 2012, Toyota Corolla Sedan 2012, Volkswagen Golf Hatchback 2012, Volkswagen Golf Hatchback 1991, Volkswagen Beetle Hatchback 2012, Volvo C30 Hatchback 2012, Volvo 240 Sedan 1993, smart fortwo Convertible 2012.
    \item \textit{country} cars: AM General Hummer SUV 2000, Aston Martin V8 Vantage Convertible 2012, BMW X5 SUV 2007, BMW X6 SUV 2012, BMW X3 SUV 2012, Buick Rainier SUV 2007, Buick Enclave SUV 2012, Cadillac SRX SUV 2012, Cadillac Escalade EXT Crew Cab 2007, Chevrolet Silverado 1500 Hybrid Crew Cab 2012, Chevrolet Traverse SUV 2012, Chevrolet HHR SS 2010, Chevrolet Tahoe Hybrid SUV 2012, Chevrolet Express Cargo Van 2007, Chevrolet Avalanche Crew Cab 2012, Chevrolet TrailBlazer SS 2009, Chevrolet Silverado 2500HD Regular Cab 2012, Chevrolet Silverado 1500 Classic Extended Cab 2007, Chevrolet Express Van 2007, Chevrolet Silverado 1500 Extended Cab 2012, Chevrolet Silverado 1500 Regular Cab 2012, Chrysler Aspen SUV 2009, Chrysler Town and Country Minivan 2012, Dodge Caravan Minivan 1997, Dodge Ram Pickup 3500 Crew Cab 2010, Dodge Ram Pickup 3500 Quad Cab 2009, Dodge Sprinter Cargo Van 2009, Dodge Journey SUV 2012, Dodge Dakota Crew Cab 2010, Dodge Dakota Club Cab 2007, Dodge Durango SUV 2012, Dodge Durango SUV 2007, Ford F-450 Super Duty Crew Cab 2012, Ford Freestar Minivan 2007, Ford Expedition EL SUV 2009, Ford Edge SUV 2012, Ford Ranger SuperCab 2011, Ford F-150 Regular Cab 2012, Ford F-150 Regular Cab 2007, Ford E-Series Wagon Van 2012, GMC Terrain SUV 2012, GMC Savana Van 2012, GMC Yukon Hybrid SUV 2012, GMC Acadia SUV 2012, GMC Canyon Extended Cab 2012, HUMMER H3T Crew Cab 2010, HUMMER H2 SUT Crew Cab 2009, Honda Odyssey Minivan 2012, Honda Odyssey Minivan 2007, Hyundai Santa Fe SUV 2012, Hyundai Tucson SUV 2012, Hyundai Veracruz SUV 2012, Infiniti QX56 SUV 2011, Isuzu Ascender SUV 2008, Jeep Patriot SUV 2012, Jeep Wrangler SUV 2012, Jeep Liberty SUV 2012, Jeep Grand Cherokee SUV 2012, Jeep Compass SUV 2012, Land Rover Range Rover SUV 2012, Land Rover LR2 SUV 2012, Mazda Tribute SUV 2011, Mercedes-Benz Sprinter Van 2012, Nissan NV Passenger Van 2012, Ram C/V Cargo Van Minivan 2012, Toyota Sequoia SUV 2012, Toyota 4Runner SUV 2012, Volvo XC90 SUV 2007.
\end{itemize}

Second, regarding the background images for the background shortcut, we use images from the Places~\cite{zhou2018IEEETrans.PatternAnal.Mach.Intell.Places} dataset, where the \textit{urban} background images are from alley, crosswalk, downtown, gas station, garage (outdoor), driveway classes, and the \textit{country} background images are forest road, field road, desert road. We use MaskFormer mentioned above to detect humans, cars, and co-occurring objects (\eg, fireplug) on Places images. Images with the aforementioned object categories detected will be filtered out. When used as the background image in UrbanCars, we resize each image to $256\times256$.

Lastly, the co-occurring objects are from LVIS~\cite{gupta2019IEEECVFConf.Comput.Vis.PatternRecognit.CVPRLVIS} based on its ground-truth instance segmentation mask, where \textit{urban} co-occurring object images are from fireplug, stop sign, street sign, parking meter, traffic light and \textit{country} co-occurring object images are from farm animals---cow, horse, sheep. We filter out instance masks with more than one connected component (\eg, instances with more than one connected component are usually occluded by other objects). When pasting to the background, the square bounding box (see above) of the co-occurring object is resized such that the side length is 25\% of the final image size.

\paragraph{Dataset Release} Since Places dataset (the source dataset for the background) does not own the copyright of images, we cannot directly release the final images in UrbanCars. Instead, we release the code that creates the UrbanCars from source datasets.

\subsection{ImageNet-Watermark (ImageNet-W) Details}
\label{appx:subsec:in_w_details}

Here we show more details about creating the ImageNet-Watermark dataset. Regarding the position, we paste the watermark at the center of the image. More specifically, the XY-position of the top-left corner of the watermark is $(0.01 W, 0.4 H)$, where $W$ and $H$ are the width and height of the input image for the models. Regarding the font size, we use 36 for the $224 \times 224$ sized images, which is the most common input size for most vision models. For large foundation models using larger input sizes, we use $62, 82, 84$  for $384 \times 384$, $512 \times 512$, $518 \times 518$ sized images, respectively, where the font sizes are approximately 0.16 times smaller to the image size. The font color for the watermark is (255, 255, 255, 128) in RGBA, which is a transparent white color. We use the open-sourced ``SourceHanSerifSC-ExtraLight''\footnote{\url{https://source.typekit.com/source-han-serif/}} as the font family.

\begin{CJK*}{UTF8}{gbsn}
\paragraph{Content of Watermark} As mentioned in \cref{subsec:imagenet_watermark}, we use ``捷径捷径捷径'' as the content of the watermark. We show the results of using other contents or languages in \cref{appx:tab:ablate_watemark_content}. When using other content in Simplified Chinese (\eg, ``一二三四五六'') or other languages (\ie, Japanese, Korean, English, and Arabic), we observe smaller IN-W Gap and Carton Gap. We conjecture this is due to the simpler shape of other contents compared to ``捷径捷径捷径'' used in the ImageNet-W. Nevertheless, the accuracy drops across different contents suggest that it is the presence of the watermark rather than its content that causes the watermark shortcut reliance. Besides, the watermark shortcut reliance is stronger when the watermark's content looks more visually similar to the pattern of the watermark in carton class images in ImageNet-1k training set, \eg, Simplified Chinese characters with complex shapes (\cf \cref{appx:fig:in1k_carton}).
\end{CJK*}

\begin{table}[h]
\centering
\begin{adjustbox}{width=0.7\linewidth}
\begin{tabular}{@{}llllll@{}}
\toprule
watermark content             & language & English translation         & Example Image & IN-W Gap & Carton Gap \\ \midrule
\begin{CJK*}{UTF8}{gbsn}捷径捷径捷径\end{CJK*}   & Simplified Chinese    & shortcut shortcut shortcut  &  \includegraphics[width=1in]{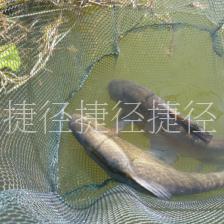}             & -26.64   & +40        \\ \midrule
\begin{CJK*}{UTF8}{gbsn}一二三四五六\end{CJK*}  & Simplified Chinese    & one two three four five six &     \includegraphics[width=1in]{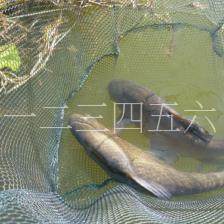}          & -6.12    & +22        \\ \midrule
\begin{CJK*}{UTF8}{goth}ショートカット\end{CJK*}  & Japanese & shortcut                    &  \includegraphics[width=1in]{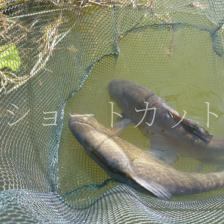}             & -2.66    & +18        \\ \midrule
\begin{CJK}{UTF8}{}\CJKfamily{mj}지름길지름길\end{CJK}   & Korean    & shortcut shortcut           &   \includegraphics[width=1in]{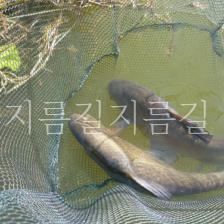}            & -12.30    & +34        \\ \midrule
shortcut  & English   & N/A                    &   \includegraphics[width=1in]{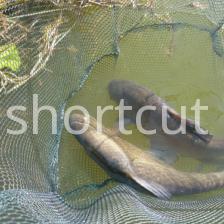}            & -6.39    & +8         \\ \midrule
abcdefghijkl & English & N/A                &   \includegraphics[width=1in]{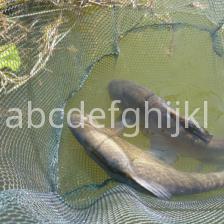}             & -5.54    & +4         \\ \midrule
\multicolumn{1}{r}{\<الاختصار>} & Arabic      & shortcut          &    \includegraphics[width=1in]{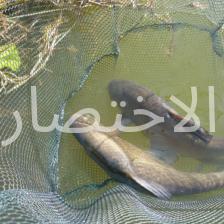}           & -7.79    & +4        \\
\bottomrule
\end{tabular}
\end{adjustbox}
\caption{Ablation study on ResNet-50's reliance on the watermark shortcut in different content and languages.
Watermarks of various contents can cause shortcut reliance.
We choose the content shown in the first row for creating ImageNet-W as it causes a larger IN-W Gap and Carton Gap and is more visually similar to the watermark that appears in the IN-1k training set.}
\label{appx:tab:ablate_watemark_content}
\end{table}

\paragraph{Dataset Release} We release the code of adding watermarks instead of directly releasing the final images. We follow AugLy~\cite{papakipos2022IEEECVFConf.Comput.Vis.PatternRecognit.CVPRWorkshopAugLy} to implement the code of adding watermarks, which is encapsulated as a function similar to PyTorch's transforms API. It is easy to use and can evaluate vision models on the fly by simply adding the watermark transform function with ImageNet-1k validation set downloaded, \ie, no need to save images with watermarks to the disk in advance.

\section{Implementation Details}

Here we present more details of the benchmark methods and our Last Layer Ensemble approach.

\subsection{Watermark Augmentation (WMK Aug)}
\label{appx:subsec:watermark_aug}

To mitigate the watermark shortcut on ImageNet, we propose simple-yet-effective watermark augmentation (WMK Aug). Concretely, we overlay a random watermark onto the training images in ImageNet-1k. The watermark is random in terms of (1) position, (2) font size, and (3) content, where we use random CJK (Chinese, Japanese, and Korean) characters in a random number of characters. The randomness of watermark augmentation in training avoids being identical to the watermark used for evaluation on ImageNet-W.

\subsection{Background Augmentation (BG Aug)}
\label{appx:subsec:bg_aug}

To mitigate the background shortcut on ImageNet, we follow \cite{xiao2021Int.Conf.Learn.Represent.Noise,ryali2021Characterizing} and use background augmentation (BG Aug). Concretely, we use unsupervised saliency segmentation developed by \citet{ryali2021Characterizing} to separate the foreground object from the backgrounds in each image. Then ``tiled'' background images are created by repeating the procedure of pasting the largest rectangular of the background onto the foreground region to cover the foreground object (more details in \cite{xiao2021Int.Conf.Learn.Represent.Noise}). Finally, to augment the background, we paste the segmented foreground object from class A onto a tiled background from class B (A$\neq$B).

\subsection{Detailed Experiment Settings}
\label{appx:subsec:detailed_exp_setting}

\paragraph{UrbanCars} On UrbanCars, we follow the standard regularization setting in \cite{sagawa2020Int.Conf.Learn.Represent.Distributionally}. Concretely, we use stochastic gradient descent (SGD) optimizer with $10^{-3}$ learning rate and $10^{-4}$ weight decay (\ie, $\ell_2$ penalty). We use 128 for the batch size. All models are trained with 300 epochs, and we use the early stopped epoch that achieves the best validation set worst-group accuracy to report the final results on the testing set. Specifically, for methods that do not use ground-truth shortcut labels (\ie, category 1, 2, 4), the worst-group accuracy is computed based on labels of both shortcuts, \ie, lowest accuracy among all eight groups. Methods using shortcut labels (\ie, category 3) may encounter the issue in which one or a subset of shortcuts remain unlabeled or even unknown. To simulate the situation, besides standard setting using labels of both shortcuts, we additionally create two settings---(1) only using BG label; (2) only using CoObj label (\cf bottom two sections in \cref{tab:urbancars_results}). In both cases, the worst-group accuracy on the validation set also only considers the label of one shortcut, \ie, the lowest accuracy among four groups based on the combination of the target label and the single shortcut label. Each experiment on UrbanCars is repeated six times using different random seeds, and we report the average results over six runs.

\paragraph{ImageNet} On ImageNet, we use last layer re-training~\cite{kirichenko2022Last} to only train the last classification layer upon a frozen feature extractor to benchmark methods in \cref{tab:imagenet_results} and our Last Layer Ensemble (LLE) method in
\cref{tab:imagenet_results_larger_arch}. Note that we directly evaluate self-supervised approaches and foundation models in \cref{tab:imagenet_results_larger_arch} without using last layer re-training. When using ResNet-50 network architecture with last layer re-training (\ie, methods in \cref{tab:imagenet_results}), we use SGD optimizer with $10^{-4}$ weight decay. For all models, we tune the learning rate over $\{10^{-2}, 10^{-3}, 10^{-4}\}$ and choose the one with the best top-1 accuracy on IN-1k. We use 1024 for the batch size. Unlike the detailed implementation in \cite{kirichenko2022Last}, we do not train the last classification layer from scratch but initialize it by the weights of ERM's last layer because we find that the latter way converges faster. Note that ERM's last layer is also re-trained (\eg, ERM in \cref{tab:imagenet_results}). When applying our LLE approach with the MAE feature extractor, we follow MAE~\cite{he2022IEEECVFConf.Comput.Vis.PatternRecognit.CVPRMasked} to use $0$ weight decay.

\subsection{Details of Benchmark Methods}
\label{appx:subsec:benchmark_methods_details}

We introduce more details (\eg, hyperparameters) of benchmark methods in each category.

\paragraph{Category 1: Standard Augmentation and Regularization} Following PyTorch's new training recipe~\cite{vryniotis2021PyTorchBlogHow}, we use $\alpha=0.2$ for Mixup, $p=0.1$ for Cutout, and $\alpha=1.0$ for CutMix on both UrbanCars and ImageNet experiments. For AugMix, we use all default hyperparameters in the original implementation. For the co-efficient of $\ell_2$ penalty of logits in SD, we use 0.1  on UrbanCars and $10^{-4}$ on ImageNet (we find that SD using $0.1$ on ImageNet achieves poor results).

\paragraph{Category 2: Targeted Augmentation for Mitigating Shortcuts} For CF+F Aug~\cite{chang2021IEEECVFConf.Comput.Vis.PatternRecognit.CVPRRobust}, based on the ground-truth masks (\cf \cref{appx:subsec:urbancars_details}), we use CF(Grey) and F(Random) for generating counterfactual and factual augmentations because they achieve the best results on Waterbirds when not using external generative models. Concretely, CF(Grey) infills the grey color to the bounding box area of the object to generate the counterfactual image, and F(Random) uses random noises to replace the background area---outside of the bounding box of the car object (more details in \cite{chang2021IEEECVFConf.Comput.Vis.PatternRecognit.CVPRRobust}).

For style transfer~\cite{geirhos2019Int.Conf.Learn.Represent.ImageNettrained} (\ie, texture augmentation or TXT Aug), we use the official code to generate Stylized ImageNet (SIN) for training. The details of BG Aug and WTM Aug are introduced in \cref{appx:subsec:bg_aug} and \cref{appx:subsec:watermark_aug}, respectively. Note that WTM Aug, TXT Aug, and BG Aug shown in \cref{tab:imagenet_results} jointly use augmented images and original IN-1k images for training.

\paragraph{Category 3: Using Shortcut Labels} We follow the original GroupDRO (gDRO)'s implementation to use 0.01 step size and $\gamma=0.1$.
For Domain Independent (DI), its number of domains is decided based on the usage of shortcut labels, \ie, 2 when using labels of only one shortcut and 4 for using labels of both shortcuts. We follow SUBG's implementation to subsample the training data to rebalance the data, where each group has (fewer but) the same number of images. For DFR, we use its DFR$^{Tr}_{Tr}$ variant where ERM's last layer is re-trained on a balanced sub-sampled training set (\ie, SUBG).

\paragraph{Category 4: Inferring Pseudo Shortcut Labels} For LfF, we follow the original implementation to set $q=0.7$. As discussed in \cref{subsec:mitigation_methods_results}, JTT and EIIL use an early-stopped ERM as the reference model to infer the pseudo shortcut labels, where we use E to denote the number of training epochs of the reference ERM model. For JTT, we use E=1 and E=2 on UrbanCars. Since JTT~\cite{liu2021Int.Conf.Mach.Learn.Just} use E=40,50,60 on Waterbirds, we also show their results on UrbanCars in \cref{appx:subsec:jtt_more_results}. We use $\lambda_\text{up}=100$ for JTT on UrbanCars. On ImageNet, we use E=1 and $\lambda_\text{up}=5$ because we found $\lambda_\text{up}=100$ (\ie, sampling wrongly predicted examples 100 times) is not scalable on the larger ImageNet dataset. For EIIL, we use E=1 and E=2 on UrbanCars and E=1 on ImageNet. We use gDRO as the invariant learner for EIIL (more details in \cite{creager2021Int.Conf.Mach.Learn.Environment}). While DebiAN uses a full network as the shortcut ``discoverer'' (more details in \cite{li2022Eur.Conf.Comput.Vis.ECCVDiscovera}), we use a single fully-connected layer on top of the feature extractor for its experiments on ImageNet under the last layer re-training setting.

\subsection{Details of Last Layer Ensemble (LLE)}
\label{appx:subsec:lle_details}

On UrbanCars, we augment background and co-occurring object visual cues to mitigate multiple shortcuts based on ground-truth masks (\cf \cref{appx:subsec:urbancars_details}).
Concretely, we use ground-truth masks of the car object and co-occurring object to (1) segment car object; (2) segment co-occurring object; (3) create the tiled background, a background-only image where the regions of the object and co-occurring object are tiled (\cf, \cref{appx:subsec:bg_aug}). To augment the background, we sample segmented car object and co-occurring object from class A and tiled background from class B (A$\neq$B), which are used to form the background-augmented images---pasting car object and co-occurring object on the tiled background. Similarly, to augment the co-occurring object, we sample the segmented car object and tiled background from class A and sample the segmented co-occurring object from class B (A$\neq$B) to create the augmented images. Note that we only use the target label of the car body type for augmentation. In other words, neither the BG shortcut labels nor the CoObj shortcut labels are used. After obtaining two types of augmented images, LLE uses three last classification layers as an ensemble---two layers for two shortcuts and one layer for the original images. The distributional shift classifier predicts three shift categories: (1) no shift (\ie, original images), (2) background shift (\ie, background-augmented images), (3) co-occurring object shift (\ie, co-occurring object augmented images).

On ImageNet, LLE uses style transfer~\cite{geirhos2019Int.Conf.Learn.Represent.ImageNettrained} (\ie, TXT Aug) to mitigate the texture shortcut, BG Aug (details in \cref{appx:subsec:bg_aug}) to mitigate the background shortcut, and WMK Aug (details in \cref{appx:subsec:watermark_aug}) to mitigate the watermark shortcut. LLE jointly trains four last classification layers as an ensemble---three layers for three shortcuts and one layer for original images in IN-1k. The distributional shift classifier predicts four categories: (1) no shift (original images from IN-1k), (2) texture shift (\ie, texture augmented images), (3) background shift (\ie, background augmented images), (4) watermark shift (\ie, watermark augmented images).

\section{Results of LfMF (Extended version of LfF)}

\begin{table}[t]
\parbox{.48\linewidth}{
\centering
\begin{adjustbox}{width=\linewidth}
\begin{tabular}{@{}lllll@{}}
\toprule
     & I.D. Acc & BG Gap         & CoObj Gap      & BG+CoObj Gap \\ \midrule
\textcolor{gray}{ERM}      & \textcolor{gray}{97.6}    & \textcolor{gray}{-15.3}         & \textcolor{gray}{-11.2}         & \textcolor{gray}{-69.2}         \\
LfMF & 97.7     & -15.6 (\textcolor{red}{$\times 1.02$} \molenohammer) & -12.7 (\textcolor{red}{$\times 1.13$} \molenohammer) & -71.2        \\ \bottomrule
\end{tabular}
\end{adjustbox}
\caption{Results of LfMF on UrbanCars dataset.}
\label{appx:tab:lfmf_urbancars}
}
\hfill
\parbox{.48\linewidth}{
\centering
\begin{adjustbox}{width=\linewidth}
\begin{tabular}{@{}lllllll@{}}
\toprule
\multicolumn{1}{c}{} & \multicolumn{1}{c|}{}      & \multicolumn{2}{c|}{watermark}                    & \multicolumn{2}{c|}{texture}            & \multicolumn{1}{c}{background} \\
                     & \multicolumn{1}{l|}{IN-1k} & IN-W Gap        & \multicolumn{1}{l|}{Carton Gap} & SIN Gap & \multicolumn{1}{l|}{IN-R Gap} & IN-9 Gap                       \\ \midrule
\textcolor{gray}{ERM}        & \textcolor{gray}{76.39}       & \textcolor{gray}{-25.40}     & \textcolor{gray}{+30}      & \textcolor{gray}{-69.43}         & \textcolor{gray}{-56.22}        & \textcolor{gray}{-5.19}                \\
LfMF                 & 76.38                      & -26.95 (\textcolor{red}{$\times 1.06$} \molenohammer) & +32 (\textcolor{red}{$\times 1.06$} \molenohammer)                     & -69.29  & -55.93                        & -5.70 (\textcolor{red}{$\times 1.10$} \molenohammer)                  \\ \bottomrule
\end{tabular}
\end{adjustbox}
\caption{Results of LfMF on ImageNet.}
\label{appx:tab:lfmf_imagenet}
}
\end{table}

One may suggest that the Whac-A-Mole problem in multi-shortcut mitigation can be solved by straightforwardly extending existing approaches designed for single-shortcut mitigation. To this end, we extend the Learning from Failure (LfF)~\cite{nam2020Adv.NeuralInf.Process.Syst.Learning} method. The original LfF method trains two networks---a bias-amplified network to identify shortcuts and a debiased network to mitigate the identified shortcuts. We extend LfF by adding the second bias-amplified network, where we investigate whether two bias-amplified networks can identify different shortcuts for mitigation. We name this method \textit{Learning from Multiple Failures} (LfMF). The results in \cref{appx:tab:lfmf_urbancars,appx:tab:lfmf_imagenet} show that LfMF still amplifies shortcuts over ERM, demonstrating that a simple extension of existing methods cannot easily solve the Whac-A-Mole problem.

\section{More Results on UrbanCars}

\subsection{More Results of JTT}
\label{appx:subsec:jtt_more_results}

In \cref{subsec:mitigation_methods_results}, we show the result of JTT when E=1 and E=2. Since JTT tunes E over $\{40, 50, 60\}$ on Waterbirds~\cite{sagawa2020Int.Conf.Learn.Represent.Distributionally} (more epochs for training the reference ERM models to infer pseudo shortcut labels). Here, we also show the results of JTT when E $\in \{40, 50, 60\}$ in \cref{appx:tab:jtt_urbancars_more_epochs}, where JTT either exhibits Whac-A-Mole results by amplifying the background shortcut or barely mitigates either shortcut compared to ERM.

\begin{table}[t]
\parbox{.48\linewidth}{
\centering
\begin{adjustbox}{width=\linewidth}
\begin{tabular}{@{}lclll@{}}
\toprule
                    & \multicolumn{1}{c|}{}                    & \multicolumn{3}{c}{shortcut reliance}               \\ \cmidrule(l){3-5}
                    & \multicolumn{1}{c|}{I.D. Acc} & BG Gap      & CoObj Gap   & BG+CoObj Gap    \\ \midrule
\textcolor{gray}{ERM}      & \textcolor{gray}{97.6}    & \textcolor{gray}{-15.3}         & \textcolor{gray}{-11.2}         & \textcolor{gray}{-69.2}         \\ \midrule
JTT (E=1)           & 95.9                                   & -8.1          & -13.3 (\textcolor{red}{$\times$1.18} \molenohammer)        & -37.6         \\
JTT (E=2)           & 94.6                                   & -23.3  (\textcolor{red}{$\times$1.52} \molenohammer)       & -5.3          & -52.1         \\
JTT (E=40)           & 97.7                                   & -15.8  (\textcolor{red}{$\times$1.03} \molenohammer)        & -10.7        & -69.3         \\
JTT (E=50)           & 97.6                                   & -14.8       & -11.0          & -67.9         \\
JTT (E=60)           & 97.2                                   & -15.1         & -10.7          & -70.5         \\
\bottomrule
\end{tabular}
\end{adjustbox}
\caption{Results of JTT when using ERM trained with other epochs (E $\in \{40, 50, 60\}$) as the reference model to infer pseudo shortcut labels on UrbanCars.}
\label{appx:tab:jtt_urbancars_more_epochs}
}
\hfill
\parbox{.48\linewidth}{
\centering
\begin{adjustbox}{width=\linewidth}
\begin{tabular}{@{}lllll@{}}
\toprule
\multicolumn{1}{c}{} & \multicolumn{1}{c|}{}         & \multicolumn{3}{c}{Shortcut Reliance}         \\ \cmidrule(l){3-5}
                     & \multicolumn{1}{c|}{I.D. Acc} & BG Gap        & CoObj Gap     & BG+CoObj Gap  \\ \midrule
\textcolor{gray}{ERM}      & \textcolor{gray}{97.6}    & \textcolor{gray}{-15.3}         & \textcolor{gray}{-11.2}         & \textcolor{gray}{-69.2}         \\ \midrule
w/o stop gradient    & 97.3                          & -3.3          & -2.6          & -7.7          \\
w/ frozen feature extractor   & 97.3                          & -11.2          & -9.0          & -43.3          \\
LLE                  & 96.7                          & -2.1 & -2.7          & -5.9          \\ \bottomrule
\end{tabular}
\end{adjustbox}
\caption{Results of ablation study of LLE on UrbanCars dataset.}
\label{appx:tab:ablate_lle_urbancars}
}
\end{table}

\subsection{Ablation Study of LLE on UrbanCars}

As mentioned in \cref{sec:our_approach_lle}, when training the distributional shift classifier, we stop the gradient from the distributional shift classifier to the feature extractor under the end-to-end training setting on UrbanCars. Here we show the ablation study in \cref{appx:tab:ablate_lle_urbancars}, where the variant without stopping the gradient achieves suboptimal results. The results demonstrate the necessity of stopping the gradient to prevent the feature extractor from learning the shortcut information used in the distributional shift classifier's supervision.

While we use the end-to-end training setting for experiments on UrbanCars, we also show the results of LLE with the last layer re-training setting (\cf frozen feature extractor in \cref{appx:tab:ablate_lle_urbancars}), which shows that using a frozen feature extractor can also improve the results over ERM, but the results are also suboptimal compared to end-to-end training.

\section{More Results on ImageNet-W}

\begin{table*}[h]
\centering
\arrayrulecolor{lightgray}
\begin{adjustbox}{width=\linewidth}
\begin{tabular}{@{}lll|cc|cc|cc@{}}
\toprule
method  & architecture & (pre)training data                  & IN-1k Acc $\uparrow$ & $P(\hat{y} = \text{carton})$ (\%) & IN-W Gap $\uparrow$ & $\Delta P(\hat{y} = \text{carton})$ (\%) $\downarrow$ & Carton Gap $\downarrow$ & $\Delta P(\hat{y} = \text{carton} \mid y = \text{carton})$ (\%) $\downarrow$ \\ \midrule
Supervised     & ResNet-50~\cite{he2016IEEEConf.Comput.Vis.PatternRecognit.CVPRDeep}    & IN-1k~\cite{deng2009IEEEConf.Comput.Vis.PatternRecognit.CVPRImageNet}                       & 76.1       & 0.07           & -26.7 & +7.56  & +40 &  +42.46    \\
MoCov3~\cite{chen2021IEEECVFInt.Conf.Comput.Vis.ICCVEmpirical} (LP)     & ResNet-50    & IN-1k                       &  74.6  & 0.08                & -20.7 & +2.94  & +44 &  +44.37    \\
Style Transfer~\cite{geirhos2019Int.Conf.Learn.Represent.ImageNettrained}  & ResNet-50    & SIN~\cite{geirhos2019Int.Conf.Learn.Represent.ImageNettrained}                       & 60.1   &   0.10             & -17.3 & +4.91  & +52 &   +50.06    \\
Mixup~\cite{zhang2018Int.Conf.Learn.Represent.mixup}     & ResNet-50    & IN-1k                       &  76.1  & 0.07                & -18.6 & +3.43  & +38 &  +39.78    \\
CutMix~\cite{yun2019IEEECVFInt.Conf.Comput.Vis.ICCVCutMix}     & ResNet-50    & IN-1k                       &  78.5  & 0.09                & -14.8 & +1.92  & +22 &  +29.61    \\
Cutout~\cite{devries2017Improved,zhong2020AAAIConf.Artif.Intell.Random}     & ResNet-50    & IN-1k                       &  77.0  & 0.08                & -18.0 & +2.93  & +32 &  +38.06    \\
AugMix~\cite{hendrycks2020Int.Conf.Learn.Represent.AugMixa}     & ResNet-50    & IN-1k                       &  77.5  & 0.09                & -16.8 & +2.61  & +36 &  +34.44    \\
BiT-M~\cite{kolesnikov2020Eur.Conf.Comput.Vis.ECCVBig}     & ResNet-50v2~\cite{he2016Eur.Conf.Comput.Vis.ECCVIdentity}    & IN-21k                       &  82.3  & 0.09                & -8.6 & +0.60  & +28 &  +29.73    \\
\midrule
Supervised     & RG-32gf      & IN-1k                       & 80.8  &   0.09                    & -14.1  &  +3.74     & +32 &     +33.43       \\
SEER~\cite{goyal2022Vision} (FT) & RG-32gf~\cite{radosavovic2020IEEECVFConf.Comput.Vis.PatternRecognit.CVPRDesigning}      & IG-1B~\cite{goyal2022Vision}  & 83.3 &    0.09                    & -6.5  &  +0.56       & +18   &  +24.26        \\
SWAG~\cite{singh2022IEEECVFConf.Comput.Vis.PatternRecognit.CVPRRevisiting} (LP) & RG-32gf      & IG-3.6B~\cite{singh2022IEEECVFConf.Comput.Vis.PatternRecognit.CVPRRevisiting}  & 84.6  &    0.08                   & -6.5  &  +0.36      & +22  &  +20.56         \\
SWAG (FT) & RG-32gf      & IG-3.6B  & 86.8  &    0.08          & -4.5 & +0.49        & +30   &  +26.03        \\ \midrule
Supervised       & ViT-B/32~\cite{dosovitskiy2021Int.Conf.Learn.Represent.Image}     & IN-1k      & 75.9     & 0.09      & -8.7         & +1.20              & +34        & +34.31                      \\
Uniform Soup~\cite{wortsman2022Int.Conf.Mach.Learn.Model} (FT) & ViT-B/32        & WIT~\cite{radford2021Int.Conf.Mach.Learn.Learning}   & 79.9       &   0.09      & -7.9  &  +0.32      & +24  &   +23.87        \\
Greedy Soup~\cite{wortsman2022Int.Conf.Mach.Learn.Model} (FT) & ViT-B/32        & WIT   & 81.0       &   0.09      & -6.5  &  +0.35      & +16  &   +23.87        \\ \midrule
Supervised     & ViT-B/16        & IN-1k                       & 81.0     &  0.08                  & -6.7 &   +0.73      & +26 & +31.28            \\
RobustViT~\cite{chefer2022Adv.NeuralInf.Process.Syst.Optimizing}     & ViT-B/16        & IN-1k                       & 80.3     &  0.08                  & -7.3 &   +0.44      & +34 & +37.06            \\
MoCov3 (LP)     & ViT-B/16        & IN-1k                       & 76.6      &   0.09                & -16.0 &  +1.97       & +22 &     +38.34       \\
MAE~\cite{he2022IEEECVFConf.Comput.Vis.PatternRecognit.CVPRMasked} (FT)  & ViT-B/16        & IN-1k & 83.7   &    0.09                  & -4.6 &  +0.67   & +24 &    +22.46        \\
SWAG (LP) & ViT-B/16        & IG-3.6B  & 81.8    &  0.08  & -7.7  &  +0.46  & +18   &   +19.74       \\
SWAG (FT) & ViT-B/16        & IG-3.6B   & 85.2       &   0.09      & -5.4  &  +0.45      & +24   &  +25.95        \\ \midrule
Supervised    & ViT-L/16        & IN-1k   & 79.6    &   0.08      & -6.2 & +0.82 & +34 & +32.57 \\
MAE (FT)    & ViT-L/16        & IN-1k                    & 85.9    &   0.09                  & -4.4 & +0.50 & +22 & +22.70 \\
SWAG (LP)    & ViT-L/16        & IG-3.6B                    & 85.1    &   0.08                  & -5.7 & +0.23 & \textbf{+6} & +9.72 \\
SWAG (FT)    & ViT-L/16        & IG-3.6B                    & 88.0    &   0.09                  & -3.2 & +0.24 & +20 & +19.14 \\
CLIP~\cite{radford2021Int.Conf.Mach.Learn.Learning} (zero-shot)   & ViT-L/14        & WIT~\cite{radford2021Int.Conf.Mach.Learn.Learning}                      & 76.5    &   0.06                  & -4.4   &  \textbf{+0.01}     & +12   & \textbf{+1.75} \\
CLIP (zero-shot)    & ViT-L/14        & LAION-400M~\cite{schuhmann2021Adv.NeuralInf.Process.Syst.WorkshopLAION400M}                      & 72.7   &    0.05                  & -4.9   &   +0.03    & +12 & +13.76  \\ \midrule
MAE (FT)    & ViT-H/14        & IN-1k                    & 86.9  &     0.08                  & -3.5 & +0.43 & +30     &  +29.59    \\
SWAG (LP)    & ViT-H/14        & IG-3.6B                    & 85.7    &   0.09                  & -4.9 & +0.19 & +8 & +12.80 \\
SWAG (FT)    & ViT-H/14        & IG-3.6B                    & 88.5    &   0.09                  & \textbf{-3.1} & +0.35 & +18 & +20.25 \\
CLIP (zero-shot)    & ViT-H/14        & LAION-2B~\cite{schuhmann2022LAION5B}                    & 77.9      &   0.06                & -3.6 & +0.03  & +16    &  +12.01       \\ \midrule
CLIP (zero-shot)   & ViT-G/14        & LAION-2B                    & 76.6      &   0.06                & -3.8 & +0.02  & +12    &  +5.61       \\
\bottomrule
\end{tabular}
\end{adjustbox}
\caption{Results of more methods (also include the methods in \cref{tab:watermark}). LP and FT stand for linear probing and fine-tuning on ImageNet-1k, respectively.}
\label{appx:tab:watermark_more_methods}
\end{table*}

\subsection{Results of More Methods on ImageNet-W}
\label{appx:subset:in_w_results_more_methods}

The results of more methods in addition to methods in \cref{tab:watermark} are shown in \cref{appx:tab:watermark_more_methods}. We observe a pervasive watermark shortcut reliance across network architecture, pretraining datasets, supervision, mitigation methods, \etc.

\subsection{ImageNetV2-W: ImageNet-W with ImageNetV2}
\label{appx:subsec:in_w_on_in_v2}

To further verify the pervasiveness of watermark shortcut reliance, we also overlay the watermark on ImageNetV2~\cite{recht2019Proc.36thInt.Conf.Mach.Learn.ImageNet} dataset to construct the ImageNet-W test set. We denote this ImageNet-W variant as \textbf{ImageNetV2-W}. The results are shown in \cref{tab:in_v2_watermark}, which is comparable to results on ImageNet-W shown in \cref{tab:watermark}. Note that some models show +0 Carton Gap results (\eg, CLIP pretrained on WIT and LAION-400M). We conjecture that it is due to the small number (\ie, ten) of carton class images in ImageNetV2. Nevertheless, they still show a considerable predicted probability increase of carton class images ($\Delta P(\hat{y} = \text{carton} \mid y = \text{carton})$). Therefore, the results on ImageNetV2-W strengthen our claim of the watermark shortcut for predicting the carton class learned by various vision models.

\begin{table*}[h]
\centering
\begin{adjustbox}{width=\linewidth}
\begin{tabular}{@{}lll|cc|cc|cc@{}}
\toprule
method  & architecture & (pre)training data                  & IN-1k Acc $\uparrow$ & $P(\hat{y} = \text{carton})$ (\%) & IN-W Gap $\uparrow$ & $\Delta P(\hat{y} = \text{carton})$ (\%) $\downarrow$ & Carton Gap $\downarrow$ & $\Delta P(\hat{y} = \text{carton} \mid y = \text{carton})$ (\%) $\downarrow$ \\ \midrule
Supervised       & ResNet-50    & IN-1k      & 63.19     & 0.09      & -26.07        & +9.29              & +70        & +53.50                      \\
MoCov3 (LP)      & ResNet-50    & IN-1k      & 61.98     & 0.09      & -19.83        & +3.33              & +40        & +44.43                      \\
Style Transfer   & ResNet-50    & SIN        & 48.63     & 0.09      & -15.88        & +5.16              & +40        & +40.28                      \\  \midrule
Supervised       & RG-32gf      & IN-1k      & 69.67     & 0.10      & -16.59        & +5.21              & +40        & +34.09                      \\
SEER (FT)        & RG-32gf      & IG-1B      & 72.48     & 0.08      & -9.00         & +0.76              & +30        & +31.03                      \\
SWAG (LP)        & RG-32gf      & IG-3.6B    & 75.51     & 0.10      & -7.48         & +0.45              & +20        & +17.57                      \\
SWAG (FT)        & RG-32gf      & IG-3.6B    & 78.18     & 0.09      & -5.67         & +0.74              & +30        & +27.15                      \\ \midrule
Supervised       & ViT-B/32     & IN-1k      & 62.99     & 0.07      & -8.45         & +1.39              & +30        & +20.97                      \\
Uniform Soup (FT)    & ViT-B/32     & WIT     & 68.58     & 0.08      & -8.57         & +0.42              & +60        & +47.84                      \\
Greedy Soup (FT)    & ViT-B/32     & WIT     & 69.54     & 0.08      & -7.43         & +0.44              & +50        & +40.78                      \\
\midrule
Supervised       & ViT-B/16     & IN-1k      & 69.55     & 0.09      & -7.55         & +0.92              & +40        & +22.66                      \\
MoCov3 (LP)      & ViT-B/16     & IN-1k      & 65.25     & 0.09      & -16.32        & +2.40              & +50        & +41.75                      \\
MAE (FT)         & ViT-B/16     & IN-1k      & 73.20     & 0.10      & -6.12         & +1.05              & +50        & +38.12                      \\
SWAG (LP)        & ViT-B/16     & IG-3.6B    & 72.87     & 0.10      & -8.66         & +0.55              & +10        & +20.01                      \\
SWAG (FT)        & ViT-B/16     & IG-3.6B    & 75.57     & 0.09      & -6.51         & +0.66              & +40        & +32.34                      \\ \midrule
Supervised       & ViT-L/16     & IN-1k      & 67.49     & 0.07      & -7.37         & +0.99              & +30        & +37.09                      \\
MAE (FT)         & ViT-L/16     & IN-1k      & 76.65     & 0.10      & -6.57         & +0.87              & +40        & +33.43                      \\
SWAG (LP)        & ViT-L/16     & IG-3.6B    & 76.64     & 0.09      & -6.71         & +0.30              & +30        & +12.46                      \\
SWAG (FT)        & ViT-L/16     & IG-3.6B    & 80.39     & 0.10      & -4.14         & +0.36              & +20        & +30.21                      \\
CLIP (zero-shot) & ViT-L/14     & WIT     & 70.87     & 0.09      & -5.29         & +0.02              & +0         & +4.20                       \\
CLIP (zero-shot) & ViT-L/14     & LAION-400M & 65.43     & 0.06      & -5.90         & +0.02              & +0         & +9.44                       \\ \midrule
MAE (FT)         & ViT-H/14     & IN-1k      & 78.46     & 0.10      & -5.26         & +0.71              & +30        & +31.43                      \\
SWAG (LP)        & ViT-H/14     & IG-3.6B    & 77.38     & 0.10      & -6.46         & +0.23              & +0         & +10.74                      \\
SWAG (FT)        & ViT-H/14     & IG-3.6B    & 81.06     & 0.09      & -4.39         & +0.46              & +10        & +21.45                      \\
CLIP (zero-shot) & ViT-H/14     & LAION-2B   & 70.92     & 0.08      & -4.44         & +0.02              & +30        & +19.09                      \\ \midrule
CLIP (zero-shot) & ViT-G/14     & LAION-2B   & 69.65     & 0.09      & -5.16         & +0.02              & +20        & +9.96                       \\ \bottomrule
\end{tabular}
\end{adjustbox}
\caption{Results of watermark shortcut with ImageNet-V2.}
\label{tab:in_v2_watermark}
\end{table*}

\clearpage

\subsection{More Qualitative Examples of Watermark Shortcut}
\label{appx:sec:watermark_qualitative_examples}

\begin{figure}[h]
    \centering
    \includegraphics[width=\linewidth]{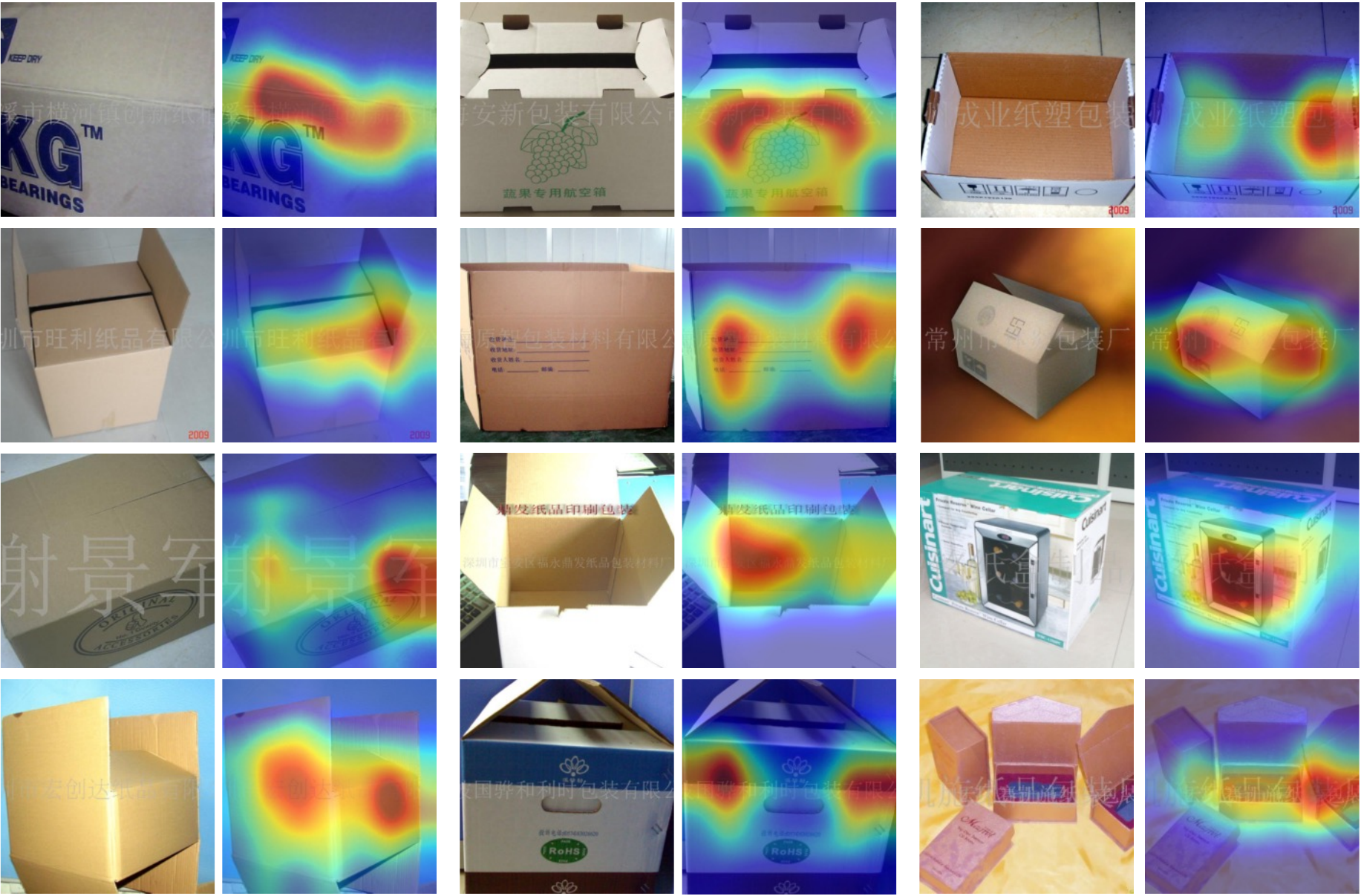}
    \caption{More examples of carton class images with watermark in ImageNet-1k training set. The saliency maps show that ResNet-50 relies on the watermark shortcut to predict carton.}
    \label{appx:fig:in1k_carton}
\end{figure}

\paragraph{Many Carton Class Images in ImageNet-1k Training Set Contain Watermark} We show more watermark examples of carton class images in ImageNet-1k training set. As shown in \cref{appx:fig:in1k_carton}, these images contain the watermark written in Chinese characters. We also show ResNet-50's saliency maps~\cite{selvaraju2017IEEEInt.Conf.Comput.Vis.ICCVGradCAM} for predicting the carton class. While they highlight the watermark region, it may still be hard to interpret because the watermark and the carton object share similar spatial locations. This could be one of the reasons why previous works did not discover this shortcut.

\begin{figure*}[h]
\centering
\begin{subfigure}{.475\textwidth}
  \centering
  \includegraphics[width=\linewidth]{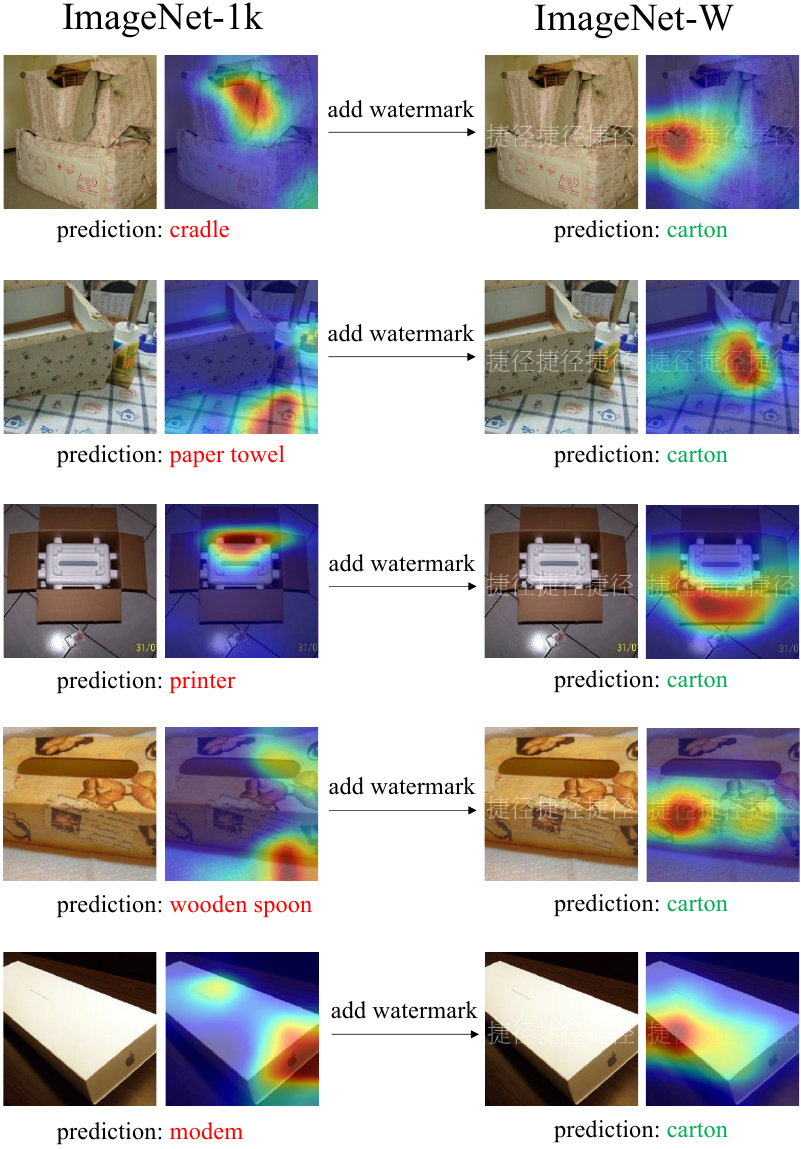}
  \caption{More examples of carton class images. ResNet-50 mispredicts many of them on ImageNet-1k validation set (left column). On ImageNet-W, after adding watermarks to carton class images from ImageNet-1k, ResNet-50 uses watermark as the shortcut to achieve correct predictions (right column).}
  \label{appx:fig:in_w_carton}
\end{subfigure}%
\hfill
\begin{subfigure}{.475\textwidth}
  \centering
  \includegraphics[width=\linewidth]{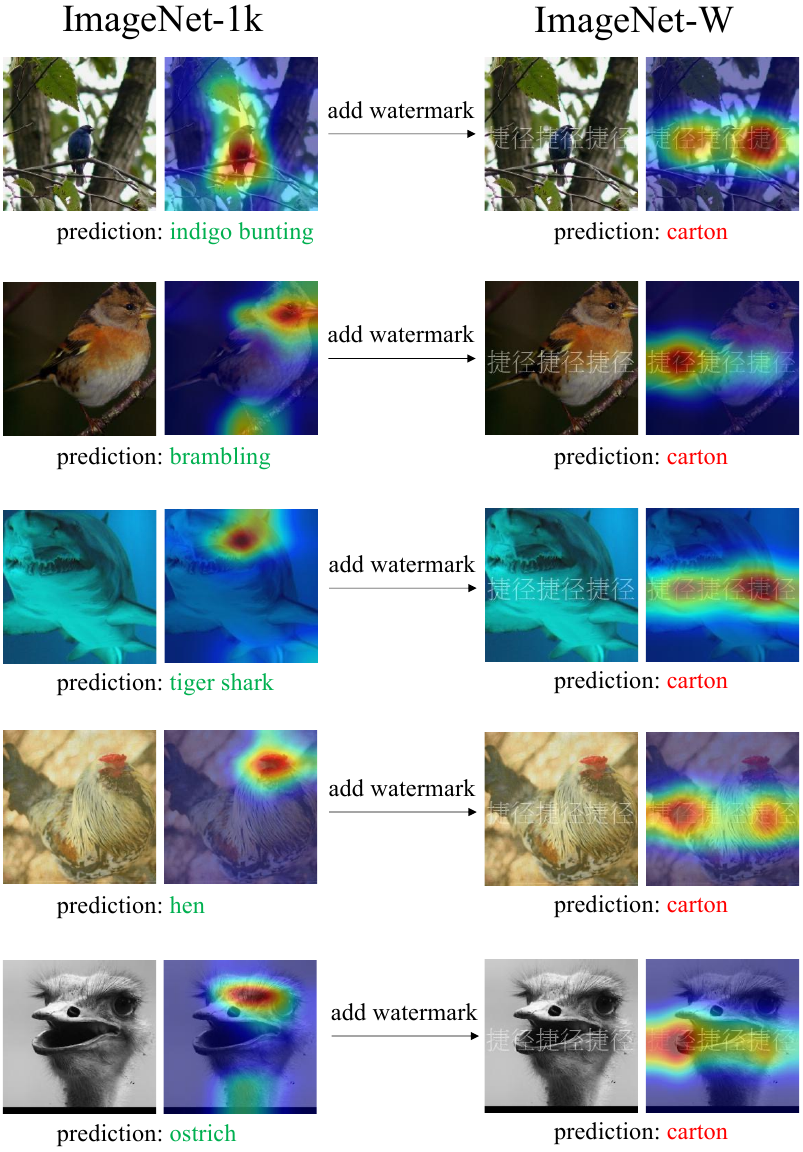}
  \caption{More examples of images that are not from the carton class. ResNet-50 predicts many of them correctly on ImageNet-1k's validation set (left column). On ImageNet-W, after adding watermarks to carton class images from ImageNet-1k, ResNet-50 uses the watermark as the shortcut to make incorrect predictions as the carton class (right column).}
  \label{appx:fig:in_w_non_carton_to_carton}
\end{subfigure}
\caption{Adding watermarks alters the prediction and focused region of ResNet-50.}
\end{figure*}

\paragraph{Adding Watermark to Carton Class Images in IN-1k Validation Set (\ie, IN-W) Leads to Carton Class Predictions} Our ImageNet-W can better address the difficulty of interpreting the watermark shortcut by providing the counterfactual explanations. In \cref{appx:fig:in_w_carton}, we first show carton class images in ImageNet-1k validation set that are predicted incorrectly by ResNet-50 (\eg, cradle, paper towel, \etc). By adding the watermark to the images, we show that not only are the predictions altered to carton but also the highlighted regions of the saliency maps are shifted to the watermark.

\paragraph{Adding Watermark to Non-Carton Class Images in IN-1k Validation Set (\ie, IN-W) Leads to Carton Class Predictions} Similarly, we also the qualitative results for non-carton class images in \cref{appx:fig:in_w_non_carton_to_carton}. While ResNet-50 makes correct predictions for non-carton class images (\eg, indigo bunting, brambling, hen, \etc) on IN-1k, the predictions are switched to carton class after adding watermarks to the images. Besides, the saliency maps show that the ResNet-50 shifts its attention from the object to the watermark shortcut.

\begin{figure}[t]
    \centering
    \includegraphics[width=\linewidth]{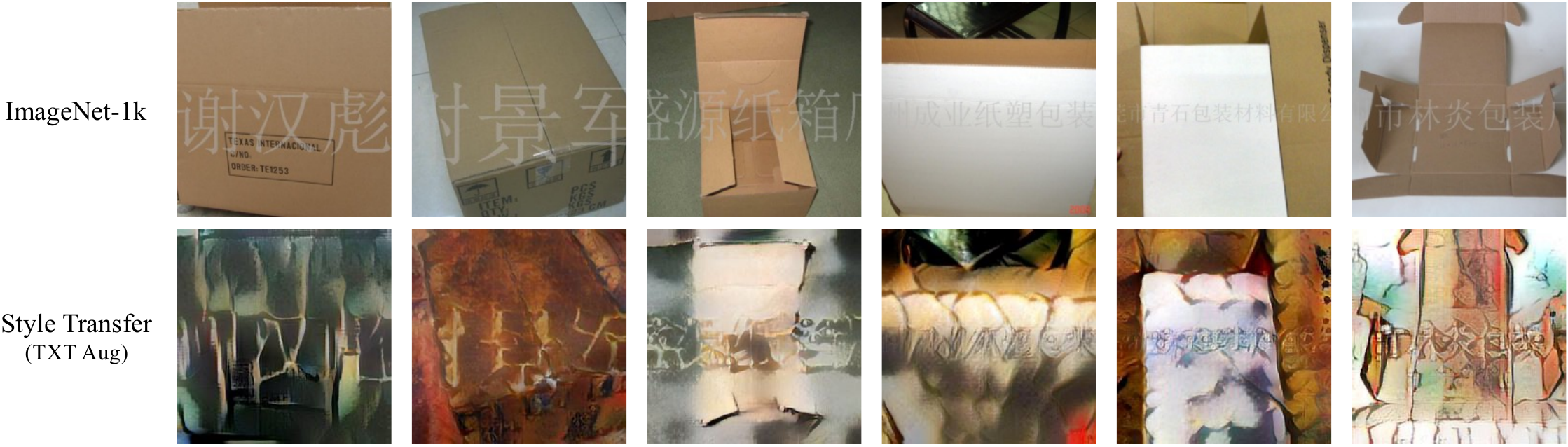}
    \caption{Examples of Style Transfer augmentation~\cite{geirhos2019Int.Conf.Learn.Represent.ImageNettrained} (TXT Aug) on carton class images from ImageNet-1k training set. Although the augmentation is designed to mitigate the texture shortcut by increasing the ``shape bias,'' it unexpectedly preserves or amplifies the shape of the watermark.}
    \label{appx:fig:style_transfer_watermark}
\end{figure}

\clearpage

\paragraph{Style Transfer (TXT Aug) Preserves or Amplifies the Shape of Watermark} In addition to \cref{fig:teaser_in_w}, we show more examples of style transfer~\cite{geirhos2019Int.Conf.Learn.Represent.ImageNettrained} augmentation for carton class images with watermark in \cref{appx:fig:style_transfer_watermark}. While the technique was originally targeted at mitigating the texture shortcut by randomizing the texture information to increase the shape bias towards the object, the shape of the watermark shortcut, as shown in \cref{appx:fig:style_transfer_watermark}, is preserved or even amplified. Watermarks in large font sizes (\cf first three images in \cref{appx:fig:style_transfer_watermark}) are still legible after style transfer. The pattern of watermarks in small font size is still retained or even more salient, \eg, the pattern of the transparent watermarks becomes more salient after style transfer when the background is white. This can explain why style transfer (\ie, TXT Aug) amplifies the watermark shortcut results in \cref{tab:imagenet_results,appx:tab:imagenet_end_to_end_results}.

\begin{figure}[t]
    \centering
    \includegraphics[width=\linewidth]{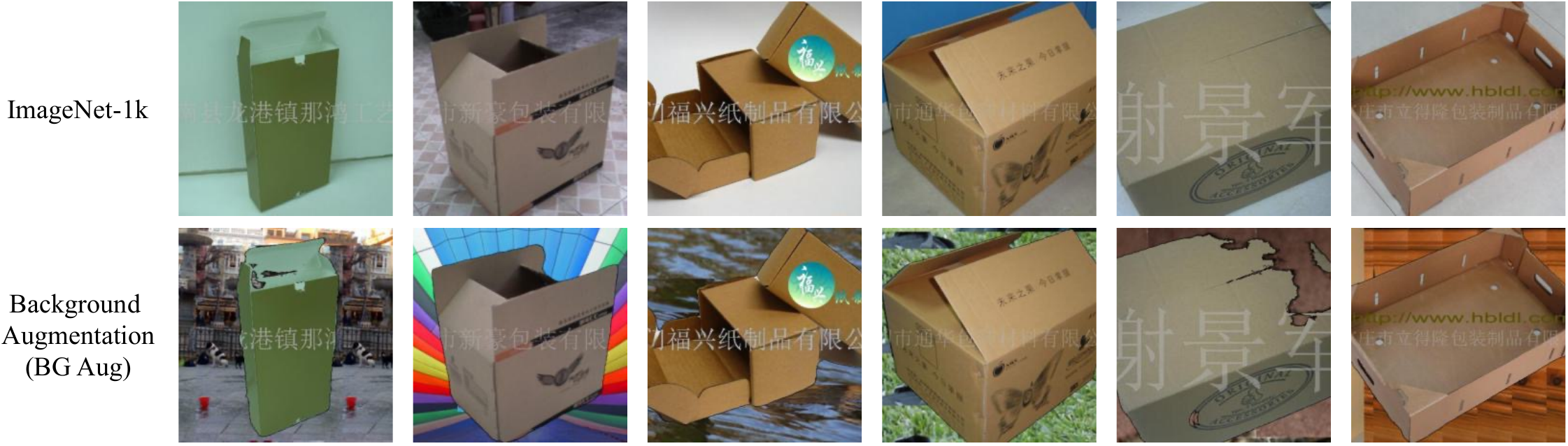}
    \caption{Examples of background augmentation (BG Aug)~\cite{xiao2021Int.Conf.Learn.Represent.Noise,ryali2021Characterizing} on carton class images from ImageNet-1k training set. BG Aug is designed to mitigate the background shortcut. However, it preserves the watermarks, leading models to pivot to the watermark shortcut.}
    \label{appx:fig:bg_aug_watermark}
\end{figure}

\paragraph{Background Augmentation (BG Aug) Preserves the Watermark Shortcut} Besides \cref{fig:teaser_in_w}, we show more examples of background augmentation (BG Aug)~\cite{xiao2021Int.Conf.Learn.Represent.Noise,ryali2021Characterizing} preserving the watermark shortcut in \cref{appx:fig:bg_aug_watermark}. Since the watermark is located over the main object, watermarks are still visible when replacing the background with a random one, which explains why BG Aug amplifies the watermark shortcut in \cref{tab:imagenet_results}. More recently, RobustViT~\cite{chefer2022Adv.NeuralInf.Process.Syst.Optimizing} uses the object mask to regularize the model to focus on the object region in the objective function, aiming to mitigate the background shortcut. Although it does not use masks to modify the input image as BG Aug does, we show that it also amplifies the watermark shortcut in \cref{appx:tab:imagenet_end_to_end_results} (\cf \cref{appx:subsec:more_existing_methods_multi_shortcut}), which can be explained by the shared spatial locations between watermark and carton object.

\section{More Results of Multi-Shortcut Mitigation on ImageNet}

\subsection{Benchmark More Existing Approaches}
\label{appx:subsec:more_existing_methods_multi_shortcut}

\paragraph{End-to-End Training} In \cref{subsec:mitigation_methods_results} and \cref{tab:imagenet_results}, we benchmark existing methods using last layer re-training~\cite{kirichenko2022Last}. Here we show the results of those methods (\ie, Mixup, Cutout, CutMix, AugMix, SD, Style Transfer, LfF, JTT, EIIL, DebiAN) using end-to-end training in \cref{appx:tab:imagenet_end_to_end_results}. We show that most of them still exhibit the Whac-A-Mole problem by achieving worse shortcut mitigation results. Although Mixup does not amplify shortcuts, its improvement over ERM is still small.

\paragraph{Big Transfer (BiT)} We also show the results of Big Transfer (BiT-M)~\cite{kolesnikov2020Eur.Conf.Comput.Vis.ECCVBig}, a foundation model pretrained on ImageNet-21k (\ie, excluding 1k classes of ImageNet-1k from the full ImageNet with 22k classes) using ResNet-50v2~\cite{he2016Eur.Conf.Comput.Vis.ECCVIdentity} architecture. \cref{appx:tab:imagenet_end_to_end_results} shows that BiT-M achieves a larger SIN Gap than ERM and barely mitigates the background shortcut.

\paragraph{RobustViT Mitigates Background Shortcut but Amplifies Other Shortcuts} RobustViT~\cite{chefer2022Adv.NeuralInf.Process.Syst.Optimizing} is a recent work designed to mitigate the background shortcut by optimizing the relevance map based on the object mask. The results in \cref{appx:tab:imagenet_end_to_end_results} show that it mitigates the background shortcut but amplifies the watermark shortcut. Besides, it also achieves a worse SIN Gap result for the texture shortcut.

\begin{table*}[h]
\centering
\begin{adjustbox}{width=\linewidth}
\begin{tabular}{@{}llllllll@{}}
\toprule
     &                    & \multicolumn{1}{l|}{}      & \multicolumn{5}{c}{shortcut reliance}                                                                                                                                 \\ \cmidrule(l){4-8}
    &         & \multicolumn{1}{l|}{}      & \multicolumn{2}{c|}{Watermark}                                     & \multicolumn{2}{c|}{Texture}                                             & Background            \\
    &                       & \multicolumn{1}{l|}{IN-1k} & IN-W Gap $\uparrow$ & \multicolumn{1}{l|}{Carton Gap $\downarrow$} & SIN Gap  $\uparrow$           & \multicolumn{1}{l|}{IN-R Gap $\uparrow$} & IN-9 Gap   $\uparrow$ \\ \midrule
\textcolor{gray}{ERM}   & ResNet-50             & \textcolor{gray}{76.13}                & \textcolor{gray}{-26.64}         & \textcolor{gray}{+40}                  & \textcolor{gray}{-69.03}               & \textcolor{gray}{-55.96}               & \textcolor{gray}{-5.53}                \\
Mixup   & ResNet-50                       & 76.11                & \textbf{-12.30}      & +38                  & -66.81               & -53.03               & \textbf{-5.06}       \\
CutMix   & ResNet-50                      & 78.58       & -19.50               & \textbf{+22}         & -72.86 (\textcolor{red}{$\times 1.06$} \molenohammer)              & -58.51 (\textcolor{red}{$\times 1.05$} \molenohammer)        & -6.25 (\textcolor{red}{$\times 1.13$} \molenohammer)               \\
Cutout   & ResNet-50                    & 77.06                & -16.29               & +32                  & -69.95 (\textcolor{red}{$\times 1.01$} \molenohammer)        & -57.32 (\textcolor{red}{$\times 1.02$} \molenohammer)              & -5.90 (\textcolor{red}{$\times 1.07$} \molenohammer)               \\
AugMix  & ResNet-50                  & 77.53                & -16.76               & +36                  & -66.38               & -51.83               & -6.42  (\textcolor{red}{$\times 1.16$} \molenohammer)              \\
SD   & ResNet-50      & 70.19                & -16.12               & +30                  & -63.63               & -59.32  (\textcolor{red}{$\times 1.06$} \molenohammer)             & -10.89 (\textcolor{red}{$\times 1.97$} \molenohammer)      \\
Style Transfer (Texture \molehammer)  & ResNet-50        & 60.18          & -17.31               & +52 (\textcolor{red}{$\times 1.30$} \molenohammer)           & \textbf{-4.32}      & \textbf{-40.76}     & -7.81 (\textcolor{red}{$\times 1.41$} \molenohammer)         \\
LfF  & ResNet-50       & 70.26                & -17.57               & +40                  & -64.34               & -56.54 (\textcolor{red}{$\times 1.01$} \molenohammer)               & -8.10 (\textcolor{red}{$\times 1.46$} \molenohammer)      \\
JTT  & ResNet-50       & 75.64                & -15.74               & +32                  & -69.04               & -55.70               & -6.75 (\textcolor{red}{$\times 1.22$} \molenohammer)      \\
EIIL  & ResNet-50       & 65.42                & -19.71               & +42 (\textcolor{red}{$\times 1.05$} \molenohammer)                 & -61.27               & -57.43 (\textcolor{red}{$\times 1.03$} \molenohammer)              & -8.66 (\textcolor{red}{$\times 1.57$} \molenohammer)      \\
DebiAN & ResNet-50        & 74.05                & -20.00               & +30                  & -67.54               & -56.70  (\textcolor{red}{$\times 1.01$} \molenohammer)             & -7.29 (\textcolor{red}{$\times 1.32$} \molenohammer)      \\ \midrule
BiT-M \scriptsize{(IN-21k)} & ResNet-50v2        & 82.32                & -8.63               & +28                  & -73.69  (\textcolor{red}{$\times 1.07$} \molenohammer)              & -51.19              & -5.25     \\ \midrule
\textcolor{gray}{ERM}   & ViT-B/16             & \textcolor{gray}{81.07}                & \textcolor{gray}{\textbf{-6.69}}         & \textcolor{gray}{\textbf{+26}}                  & \textcolor{gray}{\textbf{-62.67}}               & \textcolor{gray}{-50.36}               & \textcolor{gray}{-5.36}                \\
RobustViT (Background \molehammer) & ViT-B/16        & 80.33                & -7.35 (\textcolor{red}{$\times 1.10$} \molenohammer)              & +30  (\textcolor{red}{$\times 1.15$} \molenohammer)                & -64.06 (\textcolor{red}{$\times 1.02$} \molenohammer)             & \textbf{-45.64}              & \textbf{-5.01}     \\
\bottomrule
\end{tabular}
\end{adjustbox}
\caption{More multi-shortcut mitigation results on ImageNet. Note that methods from ERM to DebiAN use end-to-end training, which is different from the last layer re-training setting in \cref{tab:imagenet_results}. BiT-M is a foundation model pretrained on ImageNet-21k (IN-21k). RobustViT fine-tunes an ERM to mitigate the background shortcut.}
\label{appx:tab:imagenet_end_to_end_results}
\end{table*}

\subsection{Results: LLE Using Other Feature Extractors}
\label{appx:subsec:imagenet_lle_other_arch}

\begin{table*}[h]
\centering
\begin{tabular}{@{}llllllll@{}}
\toprule
      &            & \multicolumn{1}{l|}{}      & \multicolumn{5}{c}{shortcut reliance}                                                                                 \\ \cmidrule(l){4-8}
      &            & \multicolumn{1}{c|}{}      & \multicolumn{2}{c|}{Watermark}             & \multicolumn{2}{c}{Texture}             & \multicolumn{1}{|c}{Background} \\
  & train data & \multicolumn{1}{l|}{IN-1k} & IN-W Gap & \multicolumn{1}{l|}{Carton Gap} & SIN Gap & \multicolumn{1}{l|}{IN-R Gap} & IN-9 Gap                       \\ \midrule
SWAG (LP)              & IG-3.6B    & 85.74                  & -4.89           & \textbf{+8}  & -59.99          & -8.80           & -7.86                          \\
SWAG (FT)              & IG-3.6B    & 88.54         & -3.09           & +18          & -62.22          & -9.37           & -3.19                 \\
CLIP (zero-shot)       & LAION-2B   & 77.90            & -3.61           & +16          & -59.47 & \textbf{-5.61}  & -3.71                          \\
MAE (FT)               & IN-1k      & 86.89         & -3.48           & +30          & -62.29          & -33.15          & -3.24                          \\
MAE+\textbf{LLE (ours)}  & IN-1k      & 86.84                  & \textbf{-1.11}  & +28 & \textbf{-55.69} & -30.95 & \textbf{-2.35}                 \\
\bottomrule
\end{tabular}
\caption{Multi-shortcut mitigation results on ImageNet with ViT-H network architecture. LP and FT stand for linear probing and fine-tuning on ImageNet-1k, respectively. Note that there is no ERM (supervised training) available with ViT-H on ImageNet-1k.}
\label{appx:tab:imagenet_results_vit_h}
\end{table*}

We further show the results of models using the large ViT-H architecture in \cref{appx:tab:imagenet_results_vit_h}. We observed that there is no clear winner among these methods for achieving the best mitigation results on all shortcuts. Our method (LLE) can improve shortcut mitigation results over MAE in all metrics. Our method can even beat methods using extra pretraining data (\ie, SWAG and CLIP) in IN-W Gap, SIN Gap, and IN-9 Gap.

Besides, we also show the results of LLE using SWAG (FT) in ViT-B/16 architecture in \cref{appx:tab:imagenet_results_w_edge_aug}. While SWAG (LP) and SWAG (FT) suffer the Whac-A-Mole dilemma, LLE consistently mitigates multiple shortcuts jointly over ERM and SWAG (FT). Besides, we also show SWAG (FT) + LLE with edge augmentation (Edge Aug) and the results on ImageNet-Sketch. More details are introduced below (\cf \cref{appx:subsec:lle_in_sketch_results}).

\subsection{Results of LLE on ImageNet-Sketch}
\label{appx:subsec:lle_in_sketch_results}

\paragraph{Results: ImageNet-Sketch} We further show the results of LLE on ImageNet-Sketch~\cite{wang2019Adv.NeuralInf.Process.Syst.Learning} (IN-Sketch), another OOD variant of ImageNet containing sketch images in 1000 ImageNet classes. We use IN-Sketch Gap, the accuracy drop from IN-1k to IN-Sketch, to measure mitigation of color and texture shortcuts. The results in \cref{appx:tab:imagenet_results_w_edge_aug} show that our LLE method consistently improves the results over ERM, MAE, and SWAG (FT).

\begin{figure}[t]
    \centering
    \includegraphics[width=\linewidth]{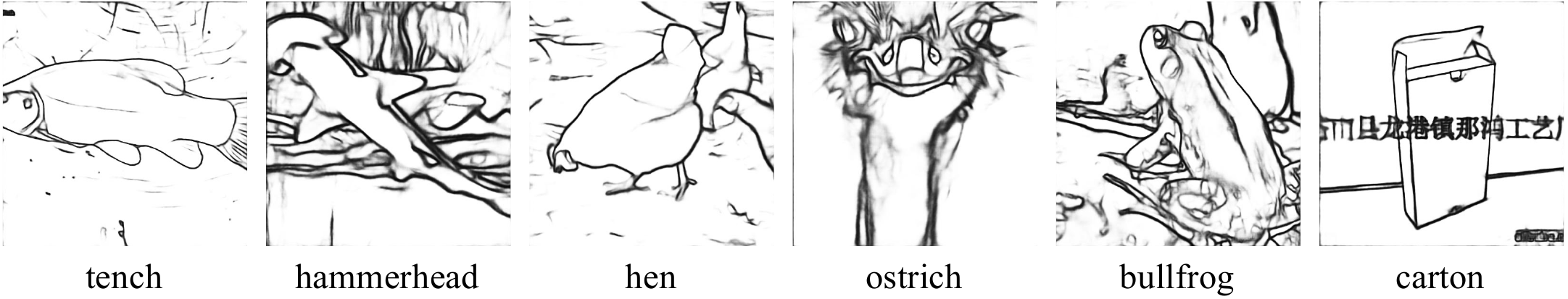}
    \caption{Example images of edge augmentation for ImageNet-1k training set to mitigate color and texture shortcuts. The ground-truth class name is shown below each image.}
    \label{appx:fig:edge_aug}
\end{figure}

\paragraph{Edge Augmentation} While style transfer augmentation could be suboptimal for mitigating the color and texture shortcuts measured by IN-Sketch, we propose edge augmentation (Edge Aug) to improve the results further. Concretely, we use \cite{poma2020IEEECVFWinterConf.Appl.Comput.Vis.WACVDense} to detect edges on images from ImageNet-1k training set. The examples are shown in \cref{appx:fig:edge_aug}, where we observe that color and texture information is successfully removed via edge detection. Similar to style transfer and background augmentation (\cf \cref{fig:teaser_in_w}), we still observe the amplified or preserved saliency of the watermark (\cf carton class image in \cref{appx:fig:edge_aug}). The edge augmentation is used to train an additional last layer in the classifier ensemble. The results in \cref{appx:tab:imagenet_results_w_edge_aug} show that using Edge Aug can further close the In-Sketch Gap and IN-R Gap---IN-R also contains sketch images. The results demonstrate the effectiveness of designing targeted augmentation to tackle the known type of shortcut.

\begin{table*}[t]
\centering
\begin{adjustbox}{width=\linewidth}
\begin{tabular}{@{}lllllllll@{}}
\toprule
       &            & \multicolumn{1}{l|}{}      & \multicolumn{5}{c}{shortcut reliance}                                                                                 \\ \cmidrule(l){4-9}
       &            & \multicolumn{1}{c|}{}      & \multicolumn{2}{c|}{Watermark}             & \multicolumn{2}{c}{Texture}             & \multicolumn{1}{|c}{Background} & \multicolumn{1}{|c}{Color and Texture} \\
 & train data & \multicolumn{1}{l|}{IN-1k} & IN-W Gap & \multicolumn{1}{l|}{Carton Gap} & SIN Gap & \multicolumn{1}{l|}{IN-R Gap} & IN-9 Gap  & IN-Sketch Gap                     \\ \midrule
\textit{arch: ResNet-50}             &      &                    &          &           &          &         &      &           \\
\textcolor{gray}{ERM}              & \textcolor{gray}{IN-1k}      & \textcolor{gray}{76.39}       & \textcolor{gray}{-25.40}           & \textcolor{gray}{+30}          & \textcolor{gray}{-69.43}          & \textcolor{gray}{-56.22}          & \textcolor{gray}{-5.19}      &   \textcolor{gray}{-52.32}         \\
\textbf{LLE (ours)}    & IN-1k      & 76.25                   & -6.18             & \textbf{+10} & \textbf{-61.02} & -54.89 & \textbf{-3.82}   &    -51.56                     \\
\textbf{LLE (ours)} + Edge Aug    & IN-1k      & 76.24                   & \textbf{-6.18}             & \textbf{+10} & -61.52 & \textbf{-53.69} & -3.95    &    \textbf{-48.25}                   \\ \midrule
\textit{arch: ViT-B/16}             &      &                    &          &           &          &         &    &             \\
\textcolor{gray}{ERM}     & \textcolor{gray}{IN-1k}     & \textcolor{gray}{81.07}          & \textcolor{gray}{-6.69}   & \textcolor{gray}{+26}          & \textcolor{gray}{-62.60}    & \textcolor{gray}{-50.36}          & \textcolor{gray}{-5.36}    &   \textcolor{gray}{-51.67}         \\ \cmidrule{2-9}
SWAG (LP)              & IG-3.6B    & 81.89                  & -7.76  \scriptsize{(\textcolor{red}{$\times 1.16$} \molenohammer)}         & +18  & -67.33 \scriptsize{(\textcolor{red}{$\times 1.08$} \molenohammer)}         & \textbf{-19.79}           & -10.39 \scriptsize{(\textcolor{red}{$\times 1.94$} \molenohammer)}       &     \textbf{-32.22}              \\
SWAG (FT)              & IG-3.6B    & 85.29         & -5.43           & +24          & -66.99  \scriptsize{(\textcolor{red}{$\times 1.07$} \molenohammer)}        & -29.55           & -4.44       &    -42.58      \\
SWAG (FT) + \textbf{LLE (ours)}  & IG-3.6B      & 85.37                  & -2.50  & \textbf{+8} & \textbf{-60.92} & -28.37 & \textbf{-3.19} & -41.52 \\
SWAG (FT) + \textbf{LLE (ours)} + Edge Aug & IG-3.6B      & 85.31                  & \textbf{-2.48}  & +12 & -61.24 & -27.78 & -3.28 & -38.37 \\ \cmidrule{2-9}
MAE (FT)  & IN-1k      & 83.72                   & -4.60             & +24 & -65.20 \scriptsize{(\textcolor{red}{$\times 1.04$} \molenohammer)} & -47.10 & -4.45  &   -47.77                      \\
MAE + \textbf{LLE (ours)}    & IN-1k      & 83.68                   & \textbf{-2.48}             & \textbf{+6} & \textbf{-58.78} & -44.96 & \textbf{-3.70}  &      -46.70                   \\
MAE + \textbf{LLE (ours)} + Edge Aug   & IN-1k      & 83.69                   & -2.54             & \textbf{+6} & -59.04 & \textbf{-43.97} & \textbf{-3.70} &    \textbf{-43.17}                      \\\midrule
\textit{arch: ViT-L/16}             &      &                    &          &           &          &         &        &         \\
\textcolor{gray}{ERM}    & \textcolor{gray}{IN-1k}      & \textcolor{gray}{79.65} & \textcolor{gray}{-6.14}   & \textcolor{gray}{+34}          & \textcolor{gray}{-61.43}           & \textcolor{gray}{-53.17}          & \textcolor{gray}{-6.50}     &  \textcolor{gray}{-52.40}          \\
MAE (FT)    & IN-1k      & 85.95         & -4.36           & +22          & -62.48 \scriptsize{(\textcolor{red}{$\times 1.02$} \molenohammer)}         & -36.46          & -3.53     &   -40.29                  \\
MAE + \textbf{LLE (ours)}  & IN-1k      & 85.84                  & \textbf{-1.74}  & \textbf{+12} & \textbf{-56.32} & -34.64  & \textbf{-2.77}   &    -39.14          \\
MAE + \textbf{LLE (ours)} + Edge Aug & IN-1k      & 85.84                  & -1.76  & +16 &  -56.52 & \textbf{-33.76}  & -2.94  &  \textbf{-36.45}             \\
\bottomrule
\end{tabular}
\end{adjustbox}
\caption{Ablation study of adding edge augmentation (Edge Aug) to LLE. Edge Aug further improves the results on ImageNet-Sketch.}
\label{appx:tab:imagenet_results_w_edge_aug}
\end{table*}

\subsection{Top-1 Accuracy of LLE on OOD Variant of ImageNet}

In this work, we mainly use the gap of accuracy between IN-1k to OOD variants of ImageNet as the metric. We also show the results of LLE in top-1 accuracy on OOD variants of ImageNet in \cref{appx:tab:top_1_acc_ood_in}, which can help future research to compare with LLE in top-1 accuracy.

Note that we do not include the top-1 accuracy on ImageNet-W. Although existing models suffer a performance drop from IN-1k to IN-W, an increased IN-W accuracy over IN-1k, which future works may achieve, also indicates the watermark shortcut reliance. Because of the counterfactual nature between IN-1k and IN-W, we encourage future works to use IN-W Gap and Carton Gap to report the watermark shortcut mitigation results, where closer to zero gaps indicate better results.

\begin{table}[h]
\centering
\begin{tabular}{@{}lllccccc@{}}
\toprule
                           &           &            & \multicolumn{1}{c|}{}      & \multicolumn{4}{c}{shortcut reliance}                                                   \\ \cmidrule(l){5-8}
                           &           &            & \multicolumn{1}{c|}{}      & \multicolumn{2}{c|}{Texture}      & \multicolumn{1}{c|}{Background} & Color and Texture \\
                           & arch      & train data & \multicolumn{1}{c|}{IN-1k} & SIN   & \multicolumn{1}{c|}{IN-R} & \multicolumn{1}{c|}{Mixed-Rand} & IN-Sketch         \\ \midrule
LLE                        & ResNet-50 & IN-1k      & 76.25                      & 15.25 & 37.31                     & 84.40                           & 24.67             \\
LLE + Edge Aug             & ResNet-50 & IN-1k      & 76.24                      & 14.72 & 38.43                     & 84.30                           & 27.99             \\ \midrule
SWAG (FT) + LLE            & ViT-B/16  & IG-3.6B    & 85.37                      & 24.45 & 68.14                     & 90.12                           & 43.85             \\
SWAG (FT) + LLE + Edge Aug & ViT-B/16  & IG-3.6B    & 85.31                      & 24.07 & 68.70                     & 89.98                           & 46.94             \\ \midrule
MAE + LLE                  & ViT-B/16  & IN-1k      & 83.68                      & 24.90 & 50.84                     & 89.41                           & 36.98             \\
MAE + LLE + Edge Aug       & ViT-B/16  & IN-1k      & 83.69                      & 24.65 & 51.85                     & 89.36                           & 40.52             \\ \midrule
MAE + LLE                  & ViT-L/16  & IN-1k      & 85.84                      & 29.52 & 62.24                     & 91.58                           & 46.70             \\
MAE + LLE + Edge Aug       & ViT-L/16  & IN-1k      & 85.84                      & 29.32 & 63.13                     & 91.41                           & 49.39             \\ \midrule
MAE + LLE                  & ViT-H/14  & IN-1k      & 86.84                      & 31.15 & 66.21                     & 93.01                           & 50.60             \\
MAE + LLE + Edge Aug                 & ViT-H/14  & IN-1k      & 86.84                      & 30.94 & 66.89                     & 92.86                           & 53.39             \\\bottomrule
\end{tabular}
\caption{Top-1 accuracy results of Last Layer Ensemble (LLE) on OOD variants of ImageNet.}
\label{appx:tab:top_1_acc_ood_in}
\end{table}

\subsection{Results of LLE on Other OOD Variants of ImageNet}
\label{appx:subsec:lle_other_ood_imagenet}

We also show the results of LLE on other OOD variants of ImageNet, including ImageNet-A~\cite{hendrycks2021IEEECVFConf.Comput.Vis.PatternRecognit.CVPRNatural} (IN-A), ImageNetV2~\cite{recht2019Proc.36thInt.Conf.Mach.Learn.ImageNet} (IN-V2), ObjectNet~\cite{barbu2019Adv.NeuralInf.Process.Syst.ObjectNet}, and ImageNet-D~\cite{rusak2022ImageNetD,rusak2021Adapting} (IN-D).
IN-D has rendition images similar to IN-R except for having additional domain annotations, \eg, clipart, infograph, \etc. Besides, IN-D also has real-domain images (\ie, IN-D real).
We report the top-1 accuracy on IN-A, IN-V2, ObjectNet, and IN-D clipart to IN-D sketch.
Regarding the types of shortcut reliance, ObjectNet measures the robustness against unusual background, viewpoint, and rotation.
The results from IN-D clipart to IN-D sketch measure the robustness against the texture shortcut.
The remaining results, \ie, IN-A, IN-V2, IN-D real, do not explicitly measure the robustness against specific shortcuts. Therefore, we denote their shortcut reliance type as ``unknown.''

The results are shown in \cref{appx:tab:lle_results_in_a_v2_objnet_d}. On both ObjectNet and IN-D datasets, LLE consistently improves the results over various baselines (\ie, ERM, SWAG (FT), and MAE (FT)) in different network architectures. When the shortcut type is unknown, LLE achieves comparable results against the baselines with slight performance improvement or drop depending on the architectures and pretraining datasets. Note that LLE is designed for mitigating multiple \textit{known} shortcuts (\cf \cref{sec:our_approach_lle}). Therefore, it may not improve the results when the types of shortcuts remain unknown. However, due to the theoretical impossibility of inferring shortcut labels~\cite{lin2022Adv.NeuralInf.Process.Syst.ZINa} and the practical difficulty of mitigating multiple unknown shortcuts, we encourage future research to tackle this problem by first interpreting the distributional shift on IN-A or IN-V2 before performing mitigation (more discussion in \cref{appx:discussion}).

\begin{table}[t]
\centering
\begin{adjustbox}{width=\linewidth}
\begin{tabular}{@{}lllllllllllll@{}}
\toprule
                           &           & \multicolumn{1}{l|}{}                   & \multicolumn{9}{c|}{shortcut reliance}                                                                                                                                                                                                     &                \\ \cmidrule(lr){4-12}
                           &           & \multicolumn{1}{l|}{}                   & \multicolumn{2}{c|}{unknown}                & \multicolumn{1}{c|}{background, viewpoint, rotation} & \multicolumn{5}{c|}{texture}                                                                         & \multicolumn{1}{c|}{unknown}   &                \\
                           & arch      & \multicolumn{1}{l|}{(pre)training data} & IN-A           & \multicolumn{1}{l|}{IN-V2} & \multicolumn{1}{l|}{ObjectNet}                       & IN-D clipart   & IN-D infograph & IN-D painting  & IN-D quickdraw & \multicolumn{1}{l|}{IN-D sketch} & \multicolumn{1}{l|}{IN-D real} & IN-D (mDE) $\downarrow$    \\ \midrule
ERM                        & ResNet-50 & IN-1k                                   & 0.02           & \textbf{63.48}             & 36.10                                                & 23.94          & 10.69          & 34.83          & 0.83           & 17.77                            & 59.86                          & 88.27          \\
LLE                        & ResNet-50 & IN-1k                                   & \textbf{0.12}  & 63.34                      & \textbf{36.67}                                       & 25.86          & \textbf{11.35} & \textbf{36.86} & 0.85           & 19.57                            & \textbf{60.60}                 & 86.79          \\
LLE + Edge Aug             & ResNet-50 & IN-1k                                   & 0.09           & 63.05                      & \textbf{36.67}                                       & \textbf{26.31} & 11.29          & 36.82          & \textbf{0.92}  & \textbf{20.72}                   & 60.57                          & \textbf{86.50} \\ \midrule
ERM                        & ViT-B/16  & IN-1k                                   & 20.88          & 69.56                      & 39.89                                                & 29.87          & 13.62          & 41.37          & 1.13           & 21.86                            & 62.75                          & 83.53          \\ \cmidrule(l){4-13}
SWAG (FT)                  & ViT-B/16  & IG-3.6B                                 & 53.01          & 75.58                      & 53.90                                                & 49.54          & 20.09          & 52.88          & 2.53           & 39.34                            & 68.17                          & 70.99          \\
SWAG (FT) + LLE            & ViT-B/16  & IG-3.6B                                 & 53.71          & \textbf{75.75}             & 54.48                                                & 51.18          & \textbf{21.63} & 54.88          & 3.19           & 41.09                            & 69.12                          & 69.25          \\
SWAG (FT) + LLE + Edge Aug & ViT-B/16  & IG-3.6B                                 & \textbf{53.75} & 75.68                      & \textbf{54.55}                                       & \textbf{51.69} & 21.43          & \textbf{54.93} & \textbf{3.59}  & \textbf{41.95}                   & \textbf{69.20}                 & \textbf{68.93} \\ \cmidrule(l){4-13}
MAE (FT)                   & ViT-B/16  & IN-1k                                   & 35.81          & \textbf{73.20}             & 47.30                                                & 34.11          & 15.27          & 44.30          & 1.17           & 27.14                            & 64.92                          & 80.15          \\
MAE (FT) + LLE             & ViT-B/16  & IN-1k                                   & 36.88          & 73.06                      & 47.63                                                & 35.25          & \textbf{16.37} & 45.90          & 1.25           & 28.66                            & 65.47                          & 78.93          \\
MAE (FT) + LLE + Edge Aug  & ViT-B/16  & IN-1k                                   & \textbf{37.00} & 72.94                      & \textbf{47.79}                                       & \textbf{35.73} & 16.10          & \textbf{45.97} & \textbf{1.34}  & \textbf{29.65}                   & \textbf{65.52}                 & \textbf{78.66} \\ \midrule
ERM                        & ViT-L/16  & IN-1k                                   & 16.64          & 67.49                      & 36.79                                                & 27.68          & 12.45          & 39.47          & 0.58           & 19.40                            & 62.04                          & 85.32          \\
MAE (FT)                   & ViT-L/16  & IN-1k                                   & \textbf{57.07} & 76.65                      & 55.31                                                & 42.64          & 18.05          & 50.14          & 3.12           & 36.87                            & 66.66                          & 74.10          \\
MAE (FT) + LLE             & ViT-L/16  & IN-1k                                   & 56.65          & \textbf{76.74}             & 55.46                                                & 43.95          & \textbf{19.31} & 51.67          & 3.27           & 38.05                            & \textbf{67.29}                 & 72.87          \\
MAE (FT) + LLE + Edge Aug  & ViT-L/16  & IN-1k                                   & 56.77          & 76.66                      & \textbf{55.65}                                       & \textbf{44.24} & 19.06          & \textbf{51.81} & \textbf{3.44}  & \textbf{38.88}                   & \textbf{67.29}                 & \textbf{72.65} \\ \midrule
MAE (FT)                   & ViT-H/14  & IN-1k                                   & 68.17          & \textbf{78.46}             & 60.47                                                & 43.69          & 19.10          & 51.29          & 3.89           & 39.17                            & 67.61                          & 72.63          \\
MAE (FT) + LLE             & ViT-H/14  & IN-1k                                   & 68.27 & 78.34                      & 60.61                                       & 45.40 & \textbf{20.80} & 52.94 & 4.24  & 40.75                   & 68.20                 & 71.12 \\
MAE (FT) + LLE + Edge Aug            & ViT-H/14  & IN-1k                                   & \textbf{68.35} & 78.32                      & \textbf{60.78}                                       & \textbf{45.76} & 20.66 & \textbf{53.06} & \textbf{4.40}  & \textbf{41.60}                   & \textbf{68.23}                 & \textbf{70.86} \\\bottomrule
\end{tabular}
\end{adjustbox}
\caption{Results of LLE on other OOD variants of ImageNet, \ie, ImageNet-A (IN-A), ImageNetV2 (IN-V2), ObjectNet, and ImageNet-D (IN-D). Except for IN-D overall results (\ie, last column), all other results are in top-1 accuracy. The overall IN-D results are reported in mDE, where lower numbers indicate better results ($\downarrow$).}
\label{appx:tab:lle_results_in_a_v2_objnet_d}
\end{table}

\section{CutMix Amplifies Background Shortcut}
\label{appx:sec:cutmix_amp_bg_shortcut}

\begin{table}[h]
\centering
\begin{tabular}{@{}lll@{}}
\toprule
       & Average Group Accuracy (\%) & Worst Group Accuracy (\%) \\ \midrule
ERM    & 87.19              & 73.88            \\
Mixup ($\alpha=0.05$)  & 87.76              & 75.73            \\
Cutout ($p=0.1$)       & 88.57              & 74.87            \\
CutMix ($\alpha=1.0$) & 74.51 \textcolor{red}{(-12.68)}         & 47.38 \textcolor{red}{(-26.50)}          \\ \bottomrule
\end{tabular}
\caption{Results of standard augmentation and regularization on Waterbirds~\cite{sagawa2020Int.Conf.Learn.Represent.Distributionally} dataset. CutMix amplifies the background shortcut on the Waterbirds dataset. $(\cdot)$: hyperparameter used in each approach.}
\label{appx:tab:cutmix_waterbirds}
\end{table}

\paragraph{Results of CutMix on Waterbirds} On UrbanCars (\cf \cref{tab:urbancars_results}) and ImageNet (\cf \cref{tab:imagenet_results,appx:tab:imagenet_end_to_end_results}), we observe that CutMix~\cite{yun2019IEEECVFInt.Conf.Comput.Vis.ICCVCutMix}  amplifies the background shortcut. We further show its background shortcut reliance on Waterbirds dataset. We use the following metrics on Waterbirds: (1) Average Group Accuracy: the unweighted average results over four groups ($\{ \text{waterbird}, \text{landbird} \} \times \{ \text{water background}, \text{land background} \}$); (2) Worst Group Accuracy: the lowest per group accuracy result. For this experiment on Waterbirds, we use the experiment setting on UrbanCars (\cf \cref{appx:subsec:detailed_exp_setting}).
\cref{appx:tab:cutmix_waterbirds} shows that CutMix achieves worse results of mitigating the background shortcut than ERM. Other techniques, \ie, Mixup and Cutout, slightly mitigates background shortcut on Waterbirds.

\paragraph{Explaining the Background Shortcut Reliance of CutMix} Since CutMix consistently amplifies the background shortcut on three datasets (\ie, UrbanCars, Waterbirds, and ImageNet), we take a closer look at its augmentation and regularization strategy. In terms of augmentation, CutMix crops a rectangular patch from one image and pastes it to the other to create the augmented image. In the regularization, the ground-truth label for the augmented image is the linear interpolation of ground-truth labels of two source images, where the interpolation co-efficient (\ie, called combination ratio $\lambda$ in CutMix) is proportional to the area of the patch. In this way, the network is regularized to predict the probability over classes that is proportional to the area in the image. Therefore, when the background takes the larger area in the image, the model predicts more on the background class instead of the smaller foreground object, leading to an amplified background shortcut reliance.

\section{Discussion}
\label{appx:discussion}

\subsection{End-to-End Training vs. Last Layer Re-Training---A Multi-Shortcut Mitigation Perspective}

Most existing shortcut mitigation methods (\eg, gDRO~\cite{sagawa2020Int.Conf.Learn.Represent.Distributionally}, SUBG~\cite{idrissi2022Conf.CausalLearn.Reason.Simple}, DI~\cite{wang2020IEEECVFConf.Comput.Vis.PatternRecognit.CVPRFairness}, JTT~\cite{liu2021Int.Conf.Mach.Learn.Just}, EIIL~\cite{creager2021Int.Conf.Mach.Learn.Environment}, LfF~\cite{nam2020Adv.NeuralInf.Process.Syst.Learning}, and DebiAN~\cite{li2022Eur.Conf.Comput.Vis.ECCVDiscovera}) train the model end-to-end. Recently, \citet{kirichenko2022Last} propose Deep Feature Reweighting (DFR), which only retrains the last classification layer of the ERM model, \ie, the feature extractor of the ERM model is frozen. DFR enjoys the advantage of efficient training compared to traditional end-to-end training approaches, which motivates us to propose our Last Layer Ensemble (LLE) method to mitigate multiple shortcuts efficiently.

However, one may worry that methods based on last layer re-training may achieve suboptimal shortcut mitigation results compared to end-to-end training approaches because the former's performance is decided by (1) how much the intended features can be extracted by the feature extractor and (2) whether the feature extractor can disentangle the intended and shortcut features. Empirically, DFR still has some gaps in combating distributional shift compared to end-to-end training methods (\eg, results of ImageNet-R and ImageNet-C in Table 3 of \cite{kirichenko2022Last}).

While \citet{kirichenko2022Last} compare the two training strategies in the single-shortcut setting, our work provides a new multi-shortcut mitigation perspective on this problem. Concretely, we compare the results of two methods---SUBG~\cite{idrissi2022Conf.CausalLearn.Reason.Simple} (\ie, an end-to-end training method) and DFR~\cite{kirichenko2022Last} (\ie, a last layer re-training method) because DFR retrains the last classification layer with SUBG method. In other words, the only difference between SUBG and DFR is the training strategy, making an apples-to-apples comparison. The results of two methods on UrbanCars in \cref{tab:urbancars_results} reveal an interesting finding. When labels of both shortcuts are used, SUBG outperforms DFR in mitigating both shortcuts. However, if labels of either shortcut are not used, SUBG amplifies the unlabeled shortcut much more significantly compared to DFR.

Therefore, from the multi-shortcut mitigation perspective, we find \textbf{last layer re-training is a more ``conservative'' strategy---although the results of mitigating the labeled shortcuts may not be optimal, it has a lower risk of significantly amplifying the unlabeled shortcuts}, which is more typical in in-the-wild datasets where types and numbers of shortcuts usually remain unknown.

\subsection{Can the problem of the watermark shortcut be addressed through data cleaning?}

We believe that using data cleaning to address the watermark shortcut problem is suboptimal for three reasons. First, it is infeasible to remove watermark images without watermark labels. Using watermark detection models may have problems because they may have shortcuts in themselves, \eg, working well for English but not Chinese watermarks. Second, removing watermarks from images (\eg, using in-painting) requires masks, which is non-trivial. Finally, removing watermark images shrinks the training set size and may amplify geographical biases. For example, we find that images with Chinese watermarks mainly from online shopping websites in China. Simply discarding these images could create performance disparity across different geographical regions~\cite{Vries_2019_CVPR_Workshops,rojas2022the}.

\subsection{Recommendation and Future Direction}

To future shortcut mitigation practitioners, we recommend the community drop the unrealistic single-shortcut assumption and be aware of the multiple-shortcut problem by having a sanity check on various inductive biases in model design, such as the usage of shortcut labels, assumption of shortcut learning during training, data augmentation, regularization, \etc.

For future shortcut mitigation dataset creators, a broader range of factors of variations (FoV) needs to be studied since some FoVs could serve as multiple shortcuts learned by models. This can be achieved by (1) manually choosing various FoVs under the controlled setting~\cite{leclerc20213DB,ibrahim2022Robustness,idrissi2022ImageNetX,barbu2019Adv.NeuralInf.Process.Syst.ObjectNet,eulig2021IEEECVFInt.Conf.Comput.Vis.ICCVDiagViB6,scimeca2022Int.Conf.Learn.Represent.Which} or (2) developing better approaches to detect and interpret shortcuts~\cite{li2021IEEECVFInt.Conf.Comput.Vis.ICCVDiscover,jain2022Distilling,eyuboglu2022Int.Conf.Learn.Represent.Domino,agarwal2022IEEECVFConf.Comput.Vis.PatternRecognit.CVPREstimating,bao2022ArXiv220413749CsLearning,deon2022ACMConf.FairnessAccount.Transpar.Spotlight,singla2021IEEECVFConf.Comput.Vis.PatternRecognit.CVPRUnderstanding} on in-the-wild datasets.

Although our work mainly focuses on the shortcut mitigation task, the importance and challenge of multiple shortcuts also apply to the shortcut detection task. For example, \citet{eyuboglu2022Int.Conf.Learn.Represent.Domino} design a shortcut detection benchmark based on CelebA, where only a single shortcut exists. Specifically, they achieve this by amplifying the correlation strength of the spurious correlation between the target attribute and the shortcut attribute. Therefore, whether or not existing shortcut detection approaches can detect multiple shortcuts is underexplored and is a promising future direction.

\subsection{Limitations}
Admittedly, our work has limitations.
For example, Last Layer Ensemble (LLE) does not address the problem of unknown types of shortcuts, which LLE may amplify. However, since mitigating unknown types of shortcuts without any inductive biases is still a theoretical~\cite{lin2022Adv.NeuralInf.Process.Syst.ZINa} and practical challenge, we advocate a human-in-the-loop solution. That is, detecting and interpreting shortcuts at the first stage. Then, LLE can be applied to mitigate the detected shortcuts.

\end{document}